\documentclass[aos,MSNbibl,citesort,dvips]{arximspdf}
\usepackage{graphicx}
\doi{10.1214/11-AOS914}
\volume{39}
\issue{5}
\pubyear{2011}
\firstpage{2686}
\lastpage{2715}

\makeatletter

\newtheorem{theorem}{Theorem}[section]
\newtheorem{lemma}[theorem]{Lemma}
\newtheorem{proposition}{Proposition}[section]

\newcommand{\reals}{\mathbb R}
\newcommand{\tr}{\operatorname{tr}}
\newcommand{\Sp}{\operatorname{Sp}}

\newcommand{\trml}{\tilde{\rmL}_2}
\newcommand{\hrmlstar}{\hat{\rmL}_2}

\newcommand{\ball}{{\mathrm{B}}}
\newcommand{\ballG}{{\mathrm{B}_\mathrm{G}}}

\newcommand{\shrtexp}{\mathbb{E}}

\newcommand{\di}{{\mathrm{d}}}
\newcommand{\eps}{\varepsilon}

\newcommand{\ang}{\operatorname{ang}}
\newcommand{\supp}{\operatorname{supp}}
\newcommand{\dist}{\operatorname{dist}}

\newcommand{\calL}{\mathcal{L}}
\newcommand{\calP}{\mathcal{P}}
\newcommand{\bbH}{\mathbb{H}}
\newcommand{\GDd}{\mathrm{G}(D,d)}
\newcommand{\bD}{\mathbf{D}}
\newcommand{\sA}{\mathrm{A}}
\newcommand{\sZ}{\mathrm{Z}}
\newcommand{\bx}{\mathbf{x}}
\newcommand{\by}{\mathbf{y}}
\newcommand{\bU}{\mathbf{U}}
\newcommand{\bV}{\mathbf{V}}
\newcommand{\bnull}{\mathbf{0}}
\newcommand{\bA}{\mathbf{A}}
\newcommand{\rmL}{\mathrm{L}}
\newcommand{\dG}{\operatorname{dist}_{\mathrm{G}}}
\newcommand{\dGK}{\operatorname{dist}_{\mathrm{G}^K}}
\newcommand{\rmF}{\mathrm{F}}
\newcommand{\rmG}{\mathrm{G}}
\newcommand{\rmX}{{\mathcal{X}}}
\newcommand{\rmY}{{\mathrm{Y}}}
\newcommand{\ddt}{\frac{\di}{\di t}}
\newcommand{\ddtp}{\frac{\di}{\di t^p}}
\newcommand{\argmin}{\arg\min}

\makeatother

\begin{document}
\begin{frontmatter}

\title{Robust recovery of multiple subspaces by geometric~{$\bolds{{l_p}}$} minimization\thanksref{T1}}
\runtitle{Robust recovery of multiple subspaces}

\thankstext{T1}{Supported in part by NSF Grants DMS-09-15064 and DMS-09-56072.}

\begin{aug}
\author[A]{\fnms{Gilad} \snm{Lerman}\corref{}\ead[label=e1]{lerman@umn.edu}}
\and
\author[A]{\fnms{Teng} \snm{Zhang}\ead[label=e2]{zhang620@umn.edu}}

\runauthor{G. Lerman and T. Zhang}

\affiliation{University of Minnesota}

\address[A]{Department of Mathematics\\
University of Minnesota\\
127 Vincent Hall\\
206 Church Street SE\\
Minneapolis, Minnesota 55455\\
USA\\
\printead{e1}\\
\hphantom{E-mail: }\printead*{e2}} 
\end{aug}

\received{\smonth{4} \syear{2011}}

%
\begin{abstract}
We assume i.i.d. data sampled from a mixture distribution with~$K$
components along fixed $d$-dimensional linear subspaces and an
additional outlier component. For $p>0$, we study the simultaneous
recovery of the $K$ fixed subspaces by minimizing the $l_p$-averaged
distances of the sampled data points from any $K$ subspaces. Under some
conditions, we show that if $0<p \leq1$, then all underlying
subspaces can be precisely recovered by $l_p$ minimization with
overwhelming probability. On the other hand, if $K>1$ and $p>1$, then
the underlying subspaces cannot be recovered or even nearly recovered
by~$l_p$ minimization. The results of this paper partially explain the successes
and failures of the basic approach of $l_p$ energy minimization for
modeling data by multiple subspaces.
\end{abstract}

%

\end{frontmatter}

\section{Introduction}
\label{secintro}

In the last decade, many algorithms have been developed to model data
by multiple subspaces.
Such hybrid linear modeling (HLM) was motivated by concrete problems
in computer vision as well as by nonlinear dimensionality reduction.
HLM is the simplest geometric framework for nonlinear dimensionality
reduction. Nevertheless, very
little theory has been developed to justify the performance of existing methods.
Here we give a rigorous analysis of the recovery of multiple subspaces
via an energy minimization.

One can model a data set $\rmX$ with $K$ subspaces obtained
by minimizing the following energy over the subspaces $\rmL_1,\ldots
,\rmL_K$:
%
%
\begin{equation}
\label{eqdeferrorksub}
e_{l_p}(\rmX,\rmL_1,\ldots,\rmL_K) = \sum_{\bx\in\rmX}
\dist^p\Biggl(\bx,\bigcup_{i=1}^K \rmL_i\Biggr),
\end{equation}
where $\dist(\cdot,\cdot)$ denotes the Euclidean distance and $p >
0$ is a fixed parameter.
For simplicity, we assume that $L_1, \ldots, L_K$ are linear
subspaces of the same dimension~$d$, and we refer to them as $d$-subspaces
(generalizations are discussed in Sections \ref{secaffinesubs}
and \ref{secmixeddimensions}).
We also assume that the data set $\rmX$ contains i.i.d. samples
from a mixture distribution $\mu$ with $K$
components along fixed $d$-subspaces and an additional outlier component.
The recovery problem asks whether with overwhelming
probability the minimization of~(\ref{eqdeferrorksub}) recovers the
underlying subspaces of $\mu$.
We show here that when $p \leq1$ the answer to this problem is
positive, whereas when $p>1$ it is negative.

Recovery problems are common in statistics, for example, recovering
a~single subspace in least squares type problems or
recovering multiple centers as in $K$-means. However, our recent
setting requires novel developments.
One issue is the strong geometric nature of our problem, resulting from
an optimization on a product space of Grassmannians.
The other is the difficulty of approximating the problem by convex
optimization (as we clarify in Section \ref{secnotconvex}).
Thus, even though it is an elementary problem in statistical learning,
it requires
the development of techniques which are currently not widely common in
statistics.

\subsection{Background and related work}
\label{subsecbackground}

Many algorithms have been developed for
HLM (see, e.g.,
\cite
{Costeira98,Torr98geometricmotion,Tipping99mixtures,Bradley00kplanes,Tseng00nearest,Kanatani01,Ho03,Vidal05,Yan06LSA,Ma07,Ma07Compression,spectralapplied,akramks08,MKFworkshop09,LBFjournal10}),
and
they find diverse applications in several areas, such as motion
segmentation in computer
vision, hybrid linear representation of images, classification of
face images and temporal segmentation of video sequences (see, e.g.,
\cite{Vidal05,Ma07,LBFjournal10}). HLM is the simplest nonlinear data
modeling and
fits within the broader frameworks of modeling data by mixture
of manifolds \cite{Arias-Castro08Surfaces} and by Whitney's stratified space~\cite{stratification10}.

The $K$-subspaces algorithm \cite
{Bradley00kplanes,Tseng00nearest,Ho03} is the most basic heuristic for
HLM, and
it suggests an iterative procedure attempting to minimize the
energy (\ref{eqdeferrorksub}) with $p=2$. It
generalizes the $K$-means algorithm, which models data by $K$ centers,
that is, $0$-dimensional affine subspaces.
Numerical experiments by Zhang et al. \cite{MKFworkshop09} have shown
that the $K$-subspaces algorithm is
in general not robust to outliers, whereas a different
method aiming to minimize (\ref{eqdeferrorksub}) with $p=1$ seems to
be robust to outliers.

There has been little investigation into performance
guarantees of the various HLM algorithms. Nevertheless, the accuracy of
segmentation under some sampling assumptions was analyzed
for two spectral-type HLM algorithms in \cite{spectraltheory}
and \cite{Arias-Castro08Surfaces}, where \cite
{Arias-Castro08Surfaces} also quantified the tolerance to outliers
(\cite{Arias-Castro08Surfaces} considers only the asymptotic case,
though applies to modeling by multiple manifolds).
For the $K$-means algorithm (which only applies to $0$-dimensional
affine subspaces), Pollard has established strong consistency \cite
{Pollard81Kmeans} and a central limit theorem \cite{Pollard82KmeansCLT}.

In \cite{lprecoverypart111}, we analyzed the $l_p$-recovery of the
``most significant''
subspace among multiple subspaces and outliers with spherically
symmetric underlying distributions.
We assume here a similar (though weaker) underlying model and rely on
some of the estimates already developed there.

\subsection{Basic conventions and notation}

We denote by $\GDd$ the Grassmannian, that is, the manifold of
$d$-subspaces of $\mathbb{R}^D$.
We measure distances between $\rmF$ and $\rmG$ in $\GDd$ by the metric
%
%
\begin{equation}
\label{eqdistgrassman}
\dG(\rmF,\rmG)=\sqrt{\sum_{i=1}^{d}\theta_i^2},
\end{equation}
where $\{\theta_i\}_{i=1}^d$ are the principal angles between $\rmF$
and $\rmG$. We use this distance since there is a simple formula for
the geodesic lines on the Grassmannian equipped with this
distance (see, e.g., \cite{lprecoverypart111}, equation 12), which is applied in
this paper.
We distinguish elements in the $K$-fold product space
$\GDd^{K}$ by the $l_\infty$ norm, that is,
%
%
\begin{equation}\label{eqdefinition}\dGK((\rmL_1,\ldots,\rmL
_K),(\hat{\rmL}_1,\ldots,\hat{\rmL}_K))=
\max_{i=1,\ldots,K}(\dG(\rmL_i,\hat{\rmL}_i)).
\end{equation}
Following \cite{Mat95}, Section 3.9, we denote by $\gamma_{D,d}$ the
``uniform'' distribution on~$\GDd$.

We denote by $a \vee b$ and $a \wedge b$ the maximum and minimum of $a$
and $b$, respectively. We designate the support
of a distribution $\mu$ by $\supp(\mu)$.
By saying ``with overwhelming probability'' or, in short, ``w.o.p.,''
we mean that the
underlying probability is at least $1-Ce^{-N/C}$, where $C$ is a
constant independent of~$N$.

\subsection{Setting of this paper}
\label{subsecprecise}

We assume an i.i.d. data set $\rmX\subseteq\reals^D$ of size~$N$
sampled from a mixture distribution representing a hybrid linear model
around distinct $d$-subspaces, $\{\rmL^*_i\}_{i=1}^K$. We in fact
consider two different types of models, but both of them have the same
basic structure.

We assume $K$ distributions, $\mu_i$, each supported on a
corresponding and distinct $d$-subspace, $\rmL^*_i$, a noise level
$\varepsilon\geq0$, and an outlier distribution, denoted by $\mu_0$.
Furthermore, for each $1\leq i\leq K$ we have a distinct noise
distribution $\nu_{i,\eps}$
with bounded support
in the  orthogonal complement $\rmL^*_i$. We assume that the $p$th
moments of $\{\|\nu_{i,\eps}\|\}_{i=1}^K$ are smaller than $\eps^p$ for
all $0<  p \leq 1$ ($p<1$ is only needed when we consider $l_p$
minimization with $p<1$).
Moreover,
if \mbox{$\varepsilon=0$}, then $\{\nu_{i,0}\}_{i=1}^K$ are the Dirac $\delta$
distributions supported on the origin within the corresponding
subspaces orthogonal to $\{\rmL^*_i\}_{i=1}^K$.

We assume\vspace*{1pt} that the underlying distributions, $\{\mu_i\}_{i=0}^K$, have
bounded supports (or possibly sub-Gaussian as explained in Section \ref
{subsecmoregeneral}).
In order to simplify our estimates, we further assume that $\supp(\mu
_i) \subseteq\ball(\bnull,1)$ for $0\leq i\leq K$.

From these pieces we construct the mixture distribution $\mu_\varepsilon$,
%
%
\begin{equation}
\label{eqmueps}
\mu_\eps=\alpha_0 \mu_0 + \sum_{i=1}^{K}\alpha_i \mu_{i} \times
\nu_{i,\eps},
\end{equation}
where
$\alpha_0 \geq0$, $\alpha_i >0$ $\forall1 \leq i \leq K$ and $\sum
_{i=0}^K \alpha_i =1$.
If $\eps=0$, then for convenience we replace the notation $\mu_\eps$
by $\mu$, that is,
%
%
\begin{equation}
\mu=\alpha_0 \mu_0 + \sum_{i=1}^{K}\alpha_i \mu_{i}.
\end{equation}

Within this basic framework, we analyze two different models. 
For $\eps\geq0$ and $\mu_\varepsilon$ as in (\ref{eqmueps}), we say
that $\mu_\eps$ is a \textit{weakly spherically symmetric HLM
distribution with noise level
$\eps$} if the $\{\mu_i\}_{i=1}^K$ are generated by rotations
(in~$\reals^D$) of a single distribution $\hat{\mu}$, such that $\hat
{\mu}(\{\bnull\})<1$,
$\supp(\hat{\mu}) \subseteq\ball(\bnull,1) \cap\hat{\rmL}$ for
some $d$-subspace\vspace*{1pt} $\hat{\rmL} \subset\reals^D$ and $\hat{\mu}$ is
spherically symmetric within $\hat{\rmL}$
(i.e., invariant to rotations within~$\hat{\rmL}$).

Our second model has weaker assumptions on the distributions of inliers
and a slightly stronger assumption on the distribution of outliers.
For $\eps\geq0$ and $\mu_\varepsilon$ as in~(\ref{eqmueps}), we say
that $\mu_\eps$ is a \textit{weak HLM
distribution with noise level $\eps$} if $\mu_i(\{\bnull\})<1$ $
\forall1 \leq i \leq K$, $\supp(\mu_\eps) \subseteq\ball(\bnull,1)$
and for some $r>0$ the uniform distribution on $\ball(\bnull,r)$ is
absolutely continuous w.r.t. the restriction of $\mu_0$ to $\ball
(\bnull,r)$.

Our theory uses the constant $\tau_0 \equiv\tau_0(d,p,\{\mu_i\}
_{i=1}^K)$. We delay its
definition to the proofs [see (\ref{eqtaudef})], but use it in the
formulation of Theorems \ref{thmhlm2} and~\ref{thmnoisyhlm2}.

\subsection{Statistical problems of this paper}
We address here two statistical problems. The simpler one is implicit
in this introduction, though clear from the proofs. It asks whether the
underlying subspaces $\{\rmL^*_i\}_{i=1}^K$ can be recovered when
$\eps= 0$ by minimizing $\shrtexp_\mu(\dist^p(\bx,\bigcup_{i=1}^K
\rmL_i))$ over $\{\rmL_i\}_{i=1}^K \subset\GDd$.
The main problem can be formulated\vspace*{1pt} using the empirical distribution
$\mu_N$ of i.i.d. sample of size $N$ from $\mu$.
It asks whether $\{\rmL^*_i\}_{i=1}^K$ can be recovered (w.o.p.) by
minimizing $\shrtexp_{\mu_N} (\dist^p(\bx,\bigcup_{i=1}^K \rmL_i))$,
which is equivalent to minimizing (\ref{eqdeferrorksub}). In the noisy
case, we extend these problems to near recovery.
When $K>1$ and $d \geq1$, these problems are nontrivial and require
complicated geometric estimates.

\subsection{Main theory}

We first formulate the exact recovery of $\{\rmL^*_i\}_{i=1}^K$ as the
unique global minimizer of the $l_p$ energy (\ref{eqdeferrorksub})
when $0 <p \leq1$.
\begin{theorem}
\label{thmhlm2}
Assume that $\mu$ is a weakly spherically symmetric HLM distribution
on $\reals^D$ without noise ($\eps=0$)
and with underlying subspaces $\{\rmL^*_i\}_{i=1}^{K} \subseteq\reals
^D$ and
mixture coefficients $\{\alpha_i\}_{i=0}^{K}$.
Let $\rmX$ be an i.i.d. data set sampled from $\mu$.
If $0 < p \leq1$ and
%
%
\begin{equation}\label{eqcondalpha2} \alpha_0< {\tau_0} \cdot
\min_{i=1,\ldots,K}\alpha_i \cdot\Bigl(1\wedge\min_{1\leq
i,j\leq K}
\dG(\rmL^*_i,\rmL^*_j)^p/2^p\Bigr),
\end{equation}
then w.o.p. the set $\{\rmL^*_1,\ldots,\rmL^*_K\}$ is the unique
global minimizer of the
ener\-gy~(\ref{eqdeferrorksub}) among all $d$-subspaces in
$\reals^D$.
\end{theorem}

Theorem \ref{thmhlm2} extends to the noisy case by allowing
near-recovery as follows
(a~counterexample for asymptotic exact recovery is shown in
Section \ref{secfapproach0}).
\begin{theorem}
\label{thmnoisyhlm2}
Assume that $\eps>0$ and $\mu_{\eps}$ is a weakly spherically
symmet\-ric HLM distribution of noise level
$\eps$ on $\reals^D$ with $K$ $d$-subspaces \mbox{$\{\rmL^*_i\}_{i=1}^K
\subseteq\reals^D$} and mixture coefficients
$\{\alpha_i\}_{i=0}^{K}$. Let $\rmX$ be an i.i.d. data sampled
from~$\mu_\eps$. If
$0<p \leq1$ and
%
%
\begin{equation}\label{eqepsbound}
\eps< 3^{-{1/p}} \Bigl(\tau_0 \cdot
\min_{i=1,\ldots,K}\alpha_i \cdot\Bigl(1\wedge\min_{1\leq
i,j\leq K}
\dG(\rmL^*_i,\rmL^*_j)^p/2^p\Bigr)-\alpha_0\Bigr)^{1/p},
\end{equation}
then any minimizer of (\ref{eqdeferrorksub}) in $\GDd^K$ has a
distance smaller than
%
%
\begin{equation}\label{eqfdefine}
f\equiv f(\varepsilon,K,d,p,\{\alpha_i\}_{i=1}^K)=
3^{1/p} \cdot\Bigl(\tau_0 \min_{1\leq j\leq
K}\alpha_j-\alpha_0\Bigr)^{-{1/p}}\cdot\eps
\end{equation}
from one of the
permutations of $(\rmL^*_1,\ldots, \rmL^*_K)$ with overwhelming probability.
\end{theorem}

At last, we formulate the impossibility to recover
$\{\rmL^*_i\}_{i=1}^K$ by $l_p$ minimization when $p>1$
(the constants $\delta_0$ and $\kappa_0$ in our formulation are
estimated in Section~\ref{secdeltakappa}).
\begin{theorem}\label{thmphasehlm2}
Assume an i.i.d. sample of $K$ $d$-subspaces $\{\rmL^*_i\}_{i=1}^{K}
\subset\GDd$
from the ``uniform'' distribution on $\GDd$, $\gamma_{D,d}$.
For $\eps\geq0$ and the sample $\{\rmL^*_i\}_{i=1}^{K}$,
let $\mu_\eps$ be a weak HLM
distribution with noise level $\eps$
and let $\rmX$ be an i.i.d. data set of size $N$ sampled from
$\mu_\eps$.
If $p> 1$ and $K>1$, then for almost every $\{\rmL^*_i\}_{i=1}^{K}$
(w.r.t. $\gamma_{D,d}^K$)
there exist positive constants $\delta_0$
and $\kappa_0$,
independent of~$N$, such that for any $\varepsilon<\delta_0$ the
minimizer of (\ref{eqdeferrorksub}),
$\hat{\rmL}_1,\ldots, \hat{\rmL}_K$, satisfies w.o.p.:
%
%
\begin{equation}
\dGK(( \hat{\rmL}_1,\ldots, \hat{\rmL}_K
),(\rmL^*_1,\ldots,\rmL^*_K))>\kappa_0 .
\end{equation}
\end{theorem}

The above theorems have direct implications for
HLM with spherically symmetric sampling along the subspaces.
Theorems \ref{thmhlm2} and \ref{thmnoisyhlm2} clarify to some extent
the robustness of two recent algorithms for HLM, which use the~$l
_1$ energy (\ref{eqdeferrorksub}): Median $K$-Flats (MKF) \cite
{MKFworkshop09} and Local Best-fit Flats (LBF) \cite{LBFcvpr10}.
Theorem \ref{thmphasehlm2} explains why common HLM strategies that use
the~$l_2$ energy (\ref{eqdeferrorksub}) (e.g., $K$-subspaces) are
generally not robust to outliers.

\subsection{Structure of the paper}
Theorems \ref{thmhlm2}, \ref{thmnoisyhlm2} and \ref{thmphasehlm2}
are proved in Sections \ref{sechlm2}, \ref{secnoisyhlm2} and \ref
{secphasehlm2}, respectively.
Section \ref{secconclusion} discusses possible extensions as well as
limitations of
our theory and suggests some open directions.


\section{\texorpdfstring{Proof of Theorem \protect\ref{thmhlm2}}{Proof of Theorem 1.1}}
\label{sechlm2}

\subsection{Preliminaries}\label{sechlm2preliminary}
We view the energy $e_{l_p}(\rmX,
\rmL_1,\ldots,\rmL_K)$ as a \mbox{function} defined on $\GDd^{K}$
while being conditioned on the fixed data set $\rmX$.
Therefo\-re, the minimizer of $e_{l_p}(\rmX,
\rmL_1,\ldots,\rmL_K)$ is an element $(\rmL'_1,\ldots,\rmL'_K)$
in~$\GDd^{K}$. Since any permutation
of its $K$ coordinates in $\GDd$ results in another minimizer, we
sometimes say that the set $\{\rmL'_1,\ldots,\rmL'_K\}$ is a
minimizer [instead of
$(\rmL'_1,\ldots,\rmL'_K)$].

We denote $e_{l_p}(\bx,\rmL_1,\ldots,\rmL_{K}) :=
e_{l_p}(\{\bx\},\rmL_1,\ldots,\rmL_{K})$ and view it as a
function on $\reals^D \times\GDd^{K}$.

We denote the set of all permutations of $(1,2,\ldots,K)$ by $\calP_K$.
We designate an open ball in $\GDd$ by $\ballG(\rmL,r)$ as opposed
to the Euclidean open ball in $\reals^D$, $\ball(\bx,r)$.

We partition $\rmX$ into the subsets $\{\rmX_i\}_{i=0}^K$ with
$\{N_i\}_{i=0}^K$ points sampled according to the distributions
$\{\mu_i\}_{i=0}^K$.

We define
%
%
\begin{equation}
\label{eqpsi}
\psi_{\mu_1}(t)={\mu_1}(\bx\in\mathbb{R}^D\dvtx-t<|\bx^T\mathbf{v}|<t),
\end{equation}
where $\mathbf{v}$ is an arbitrarily fixed unit vector in $\rmL_1^*$
[due to the spherical symmetry of~$\mu_1$ within $\rmL_1^*$, (\ref{eqpsi})
is independent of $\mathbf{v}$].
We note that since $\{\mu_i\}_{i=1}^K$ are generated by a single distribution,
$\psi_{\mu_1}(t) = \psi_{\mu_i}(t)$ $ \forall 2 \leq i \leq K$.
The invertibility of $\psi_{\mu_1}$ is established in
\cite{lprecoverypart111}, Appendix
A.2, and an estimate of~$\psi_{\mu_1}$ for a
uniform distribution on a $d$-dimensional ball appears in
\cite{lprecoverypart111}, Appendix~A.1.

Theorem \ref{thmhlm2} uses the constant $\tau_0$, which we can now
define as follows:
%
%
\begin{equation}\label{eqtaudef}
\tau_0:=\frac{(1-\mu_1(\{\bnull\})) \cdot2^{p-1}\cdot\psi_{\mu
_1}^{-1} ( ({1+(2K-1)\mu_1(\{\bnull\})})/({2K})
)^p}{(\pi\sqrt{d})^p }.
\end{equation}

In the special case where $\mu_1$ is the uniform distribution on
$\ball(\b0,1)\cap\rmL_1$, then the estimate of $\psi_{\mu}$
in \cite{lprecoverypart111}, Section A.1, implies the following lower
bound for
$\tau_0$:
\[
\tau_0>  \frac{1}{2^{p+1} \cdot K^p \cdot d^{3p/2}}.
\]
Consequently, Theorem \ref{thmhlm2} holds in this case if $\tau_0$ in
(\ref{eqcondalpha2}) is replaced by $1/(2^{p+1} \cdot K^p \cdot
d^{3p/2})$. Furthermore, it follows from basic scaling arguments that
if $\mu_1$ is the uniform distribution on $\ball(\b0,r_1)\cap\rmL_1$
and $\supp(\mu_0)\subseteq \ball(\b0,r_2)$, where $r_1$ and $r_2$ are
any positive numbers, then
\[
\tau_0>  \frac{r_1^p}{2^{p+1} \cdot K^p \cdot d^{3p/2} \cdot r_2^p}.
\]

\subsection{Auxiliary lemmata}\label{sechlm2lemma}

The following lemmata are used throughout this proof (Lemma \ref
{lemmaelldistest} is proved in the \hyperref[app]{Appendix} and Lemma
\ref
{lemmadistance} in \cite{lprecoverypart111}, Appendix~A.2).
\begin{lemma}\label{lemmaelldistest}
$\!\!\!$Suppose that\vspace*{1pt}
$\rmL_1,\hat{\rmL}_1,\ldots,\hat{\rmL}_K\,{\in}\,\GDd, p\,{>}\,0$
and $\mu_1$~is~a~sphe\-rically symmetric distribution in
$\ball(\mathbf{0},1)\cap\rmL_1$. If $\min_{1\leq j \leq
K}\dG(\rmL_1,\hat{\rmL}_j)>\eps$, then
\[
\shrtexp_{{\mu}_1}(e_{l_p}(\bx,\hat{\rmL}_1,\ldots,
\hat{\rmL}_K))
>\tau_0 \eps^p.
\]
\end{lemma}
\begin{lemma}\label{lemmadistance}
For any $\bx\in\reals^D$ and $\rmL_1,\rmL_2\in\GDd$,
\[
|{\dist}(\bx,\rmL_1)-\dist(\bx,\rmL_2)|\leq\|\bx\| \dG(\rmL
_1,\rmL_2).
\]
\end{lemma}

\subsection{Proof in expectation}

We verify Theorem \ref{thmhlm2} ``in expectation,'' whe\-reas later
sections extend the proof to hold w.o.p.
We use\vspace*{1pt} the following notation w.r.t. the fixed $d$-subspaces $\rmL^*_1$,
$\rmL^*_2,
\ldots, \rmL^*_K$, $\hat{\rmL}_1$, $\hat{\rmL}_2, \ldots,
\hat{\rmL}_K\in\GDd$:
%
%
\begin{equation}\label{eqdefIi} I(i)=\mathop{\argmin}_{1\leq j\leq
K}\dG(\rmL^*_i,\hat{\rmL}_j) \qquad\forall 1\leq i \leq K
\end{equation}
and
%
%
\begin{equation}\label{eqdefd0} d_0= \min_{i_1,i_2,\ldots,i_K \in
\calP_K}\dGK((\rmL^*_{i_1},\ldots,\rmL^*_{i_K}),(\hat{\rmL
}_1,\ldots,\hat{\rmL}_K)).
\end{equation}

The ``expected version'' of Theorem \ref{thmhlm2} is formulated and
proved as follows.
\begin{proposition}\label{propnonneighbor}
Suppose that $\hat{\rmL}_1,\ldots,\hat{\rmL}_K$ are arbitrary
subspaces in $\GDd$, $0<p\leq1$, and $I$ is defined w.r.t. $\{\hat
{\rmL}_i\}_{i=1}^K$
and the underlying subspaces $\{\rmL^*_i\}_{i=1}^K$. If $(I(1),\ldots
,I(K))$ is a permutation of
$(1,\ldots,K)$, then
%
%
\begin{eqnarray}\label{eqforpermutation}
&&\shrtexp_{\mu} e_{{l_p}}(\bx, \hat{\rmL}_1,\ldots,\hat{\rmL}_K)-
\shrtexp_{\mu} e_{{l_p}}(\bx,
\rmL^*_1,\ldots,\rmL^*_K)\nonumber\\[-8pt]\\[-8pt]
&&\qquad\geq\Bigl(\tau_0\min_{1\leq j\leq K}\alpha_j-\alpha_0\Bigr)
d_0^p.\nonumber
\end{eqnarray}
On the other hand, if $(I(1),\ldots,I(K))$ is not a permutation of
$(1,\ldots,K)$, then
%
%
\begin{eqnarray}
&&\shrtexp_{\mu} e_{{l_p}}(\bx, \hat{\rmL
}_1,\ldots,\hat{\rmL}_K)-
\shrtexp_{\mu} e_{{l_p}}(\bx, \rmL^*_1,\ldots,\rmL^*_K)
\nonumber\\[-8pt]\\[-8pt]
&&\qquad\geq\tau_0
\Bigl(\min_{1\leq j\leq K}\alpha_j\Bigr) \Bigl(\min_{1\leq
i,j\leq
K}\dG^p(\rmL^*_i,\rmL^*_j)/2\Bigr)-\alpha_0.\nonumber
\end{eqnarray}
\end{proposition}
\begin{pf}
We
define
\[
M=\mathop{\arg\max}_{1\leq i\leq
K}\dG\bigl(\rmL^*_{i},\hat{\rmL}_{I(i)}\bigr).\vadjust{\goodbreak}
\]
Assume first that $(I(1),\ldots,I(K))$ is a permutation of
$(1,\ldots,K)$. Using the
definition of $I$, we have
%
%
\begin{eqnarray} \label{eqlowboundbyd}
\min_{1\leq j\leq
K}\dG(\rmL^*_{M},\hat{\rmL}_j)&=& \dG\bigl(\rmL^*_{M},\hat{\rmL
}_{I({M})}\bigr)\nonumber\\
&=&
\dGK\bigl((\rmL^*_1,\ldots,\rmL^*_K),\bigl(\hat{\rmL}_{I(1)},\ldots,\hat
{\rmL}_{I(K)}\bigr)\bigr)\\
&=&
d_0.\nonumber
\end{eqnarray}
Combining (\ref{eqlowboundbyd}) with
Lemma \ref{lemmaelldistest}, we obtain that
%
%
\begin{eqnarray}\label{eqX1permutation}
&&\shrtexp_{\mu_{M}} e_{{l
_p}}(\bx, \hat{\rmL}_1,\ldots,\hat{\rmL}_K)-
\shrtexp_{\mu_{M}} e_{{l_p}}(\bx,
\rmL^*_1,\ldots,\rmL^*_K)\nonumber\\[-8pt]\\[-8pt]
&&\qquad=\shrtexp_{\mu_{M}}
e_{{l_p}}(\bx,
\hat{\rmL}_1,\ldots,\hat{\rmL}_K) > \tau_0 d_0^p.\nonumber
\end{eqnarray}
For any $\bx\in\rmX_0$, let
$m(\bx)=\argmin_{1\leq i\leq K}\dist(\bx,\rmL^*_i)$,
$\hat{m}(\bx)=\break\argmin_{1\leq i\leq
K}\dist(\bx,\hat{\rmL}_i)$ and note that
%
%
\begin{eqnarray}\label{eqX0permu}
&&e_{l_p}(\bx,
\hat{\rmL}_1,\ldots,\hat{\rmL}_K)-e_{l_p}(\bx,
\rmL^*_1,\ldots,\rmL^*_K)\nonumber\\
&&\qquad=\dist\bigl(\bx,\hat{\rmL}_{\hat{m}(\bx
)}\bigr)^p
-\dist\bigl(\bx,\rmL^*_{m(\bx)}\bigr)^p\nonumber\\
&&\qquad\geq
\dist\bigl(\bx,\hat{\rmL}_{\hat{m}(\bx)}\bigr)^p-\dist\bigl(\bx,\rmL
^*_{I^{-1}(\hat{m}(\bx))}\bigr)^p\\
&&\qquad\geq
-\|\bx\|^p\dG\bigl(\hat{\rmL}_{\hat{m}(\bx)},\rmL^*_{I^{-1}(\hat
{m}(\bx))}\bigr)^p\nonumber\\
&&\qquad\geq
-\|\bx\|^pd_0^p \geq-d_0^p,\nonumber
\end{eqnarray}
where the second inequality in (\ref{eqX0permu}) uses
Lemma \ref{lemmadistance}. Therefore,
%
%
\begin{equation}\label{eqX0permutation}
\shrtexp_{\mu_0} e_{{l_p}}(\bx, \hat{\rmL
}_1,\ldots,\hat{\rmL}_K)-
\shrtexp_{\mu_0} e_{{l_p}}(\bx,
\rmL^*_1,\ldots,\rmL^*_K)>-d_0^p.
\end{equation}
At last, we observe that
%
%
\begin{eqnarray}\label{eqellcombine}
&& \shrtexp_{\mu} e_{{l_p}}(\bx,
\hat{\rmL}_1,\ldots,\hat{\rmL}_K)- \shrtexp_{\mu}
e_{{l_p}}(\bx, \rmL^*_1,\ldots,\rmL^*_K) \nonumber\\
&&\qquad\geq\alpha_{M} \bigl(\shrtexp_{\mu_{M}}e_{{l_p}}(\bx,
\hat{\rmL}_1,\ldots,\hat{\rmL}_K) -\shrtexp_{\mu_{M}}
e_{{l_p}}(\bx, \rmL^*_1,\ldots,\rmL^*_K)\bigr)
\\
&&\qquad\quad{}+\alpha_0\bigl(\shrtexp_{\mu_0}e_{{l_p}}(\bx,
\hat{\rmL}_1,\ldots,\hat{\rmL}_K)-\shrtexp_{\mu_0}
e_{{l_p}}(\bx,
\rmL^*_1,\ldots,\rmL^*_K)\bigr).\nonumber
\end{eqnarray}
The proposition in this case thus follows from (\ref
{eqX1permutation}), (\ref{eqX0permutation}) and
(\ref{eqellcombine}).

Next, we assume that $I(1),\ldots,I(K)$ is not a permutation of
$1,2,\ldots,K$.
In this case, there exist $1\leq n_1,n_2 \leq K$ such that
$I(n_1)=I(n_2)$ and, consequently,
%
%
\begin{eqnarray}\label{eqdistestnonpermu}
2 \min_{1\leq j\leq K} \dG(\rmL^*_{M},\hat
{\rmL}_j)&=&
2\dG\bigl(\rmL^*_{M},\hat{\rmL}_{I({M})} \bigr)\nonumber\\
&\geq&\dG\bigl(\rmL^*_{n_1},\hat{\rmL}_{I(n_1)} \bigr)
+\dG\bigl(\rmL
^*_{n_2},\hat{\rmL}_{I(n_2)} \bigr)\nonumber\\[-8pt]\\[-8pt]
&\geq&\dG(\rmL^*_{n_1},\rmL^*_{n_2}) \nonumber\\
&\geq&\min_{1\leq i,j\leq
K}\dG(\rmL^*_i,\rmL^*_j).\nonumber
\end{eqnarray}
Combining (\ref{eqdistestnonpermu}) and
Lemma \ref{lemmaelldistest} [applied with $\eps=\min_{1\leq
i,j\leq K}\dG(\rmL^*_i,\break\rmL^*_j)/2$], we obtain that
%
%
\begin{eqnarray}\label{eqX1nonpermutation}
&&\shrtexp_{\mu_{M}}
e_{{l_p}}(\bx, \hat{\rmL}_1,\ldots,\hat{\rmL}_K)-
\shrtexp_{\mu_{M}} e_{{l_p}}(\bx,
\rmL^*_1,\ldots,\rmL^*_K)\nonumber\\[-8pt]\\[-8pt]
&&\qquad> \tau_0
\Bigl(\min_{1\leq i,j\leq K}\dG(\rmL^*_i,\rmL^*_j)/2\Bigr)^p.\nonumber
\end{eqnarray}
Finally, since the support of $\mu_0$ is contained in $\ball(\bnull
,1)$, we note that
%
%
\begin{equation}\label{eqX0nonpermutation}
\shrtexp_{\mu_0} e_{{l_p}}(\bx, \hat{\rmL
}_1,\ldots,\hat{\rmL}_K)-
\shrtexp_{\mu_0} e_{{l_p}}(\bx, \rmL^*_1,\ldots,\rmL^*_K)
\geq-1.
\end{equation}
The proposition is thus concluded from (\ref{eqellcombine}), (\ref
{eqX1nonpermutation}) and
(\ref{eqX0nonpermutation}).
\end{pf}

\subsection{Proof in a local ball by calculus on the Grassmannian}
\label{seclocalproof}

We cannot directly extend (\ref{eqforpermutation}) to an estimate w.o.p.,
since its lower bound is a multiplication of $d_0^p$, which approaches
zero as the set $\{\rmL_i\}_{i=1}^K$ approaches $\{\rmL^*_i\}_{i=1}^K$.
We will need to exclude a ball in $\GDd^K$ around $\{\rmL^*_i\}
_{i=1}^K$ before such an extension. We thus prove here
that $\{\rmL^*_i\}_{i=1}^K$ is a unique global minimizer w.o.p. in a
local ball. In Section \ref{sechlm2global} we extend Proposition \ref
{propnonneighbor} to an estimate w.o.p. outside this ball and conclude
the theorem.

We show that there exists a sufficiently small number $\gamma_1$
such that $\{\rmL^*_i\}_{i=1}^K$ is the unique global minimizer
w.o.p. of $e_{l_p}$ in $\ballG((\rmL^*_{i_1},\ldots,\rmL
^*_{i_K}),\gamma_1)$.
Since~$e_{l_p}$ is permutation invariant, it is also the unique
global minimizer in
\[
\bigcup_{i_1,i_2,\ldots,i_K \in
\calP_K}\ballG((\rmL^*_{i_1}, \ldots,\rmL^*_{i_K}),\gamma_1).
\]

In order to simplify notation in this part of the proof, we will adopt
WLOG the
convention that the RHS of (\ref{eqdefinition}) occurs at $i=1$, that is,
%
%
\begin{equation}
\label{eqassume1}
\dG(\rmL^*_1,\hat{\rmL}_1) = \max_{i=1,\ldots,K}(\dG(\rmL
^*_i,\hat{\rmL}_i)).
\end{equation}

Following this convention and the fact that $e_{l_p}(\sum
_{i=2}^{K}\rmX_i,
\rmL^*_1,\ldots,\rmL^*_K)=0$, it is enough to prove that $(\rmL
^*_1,\ldots,\rmL^*_K)$ is the unique global minimizer
w.o.p. of $e_{l_p}(\rmX_0\cup\rmX_1, \rmL_1, \ldots, \rmL_K)$ in
$\ballG((\rmL^*_1,\ldots, \rmL^*_K),\gamma_1)$, for
sufficiently small $\gamma_1$.

Let $t_0 :=\dG(\rmL^*_1,\hat{\rmL}_1)$.
For each $1\leq i\leq K$, we parametrize according to arc length the
geodesic lines from
$\rmL^*_i$ to $\hat{\rmL}_i$ by functions $\rmL_i(t)$, $1\leq i\leq
K$, on the interval
$[0,t_0]$ such that
%
%
\begin{equation}
\label{eqcondparam}
\rmL_i(0)=\rmL^*_i \quad\mbox{and}\quad \rmL_i(t_0)=\hat{\rmL}_i.
\end{equation}
We will prove that for sufficiently small $\gamma_1>0$,
%
%
\begin{equation}\label{eqlocal7}\quad
\ddtp\bigl(e_{l_p}\bigl(\rmX_0\cup\rmX_1,
\rmL_1(t),\ldots,\rmL_K(t)\bigr)\bigr)>0 \qquad\mbox{for all
$0\leq t\leq\gamma_1$ w.o.p.}
\end{equation}
This will clearly imply our desired result.

Our proof of (\ref{eqlocal7}) is based on the following estimate:
%
%
\begin{equation}\label{eqderivlowbound}
\ddtp(e_{l_p}(\bx,
\rmL_1(t),\ldots,\rmL_K(t)))\bigg|_{t=0} \geq-\|\bx\|.\vadjust{\goodbreak}
\end{equation}
In order to establish (\ref{eqderivlowbound}), we denote
$j=\argmin_{1\leq i\leq K}\dist(\bx,\rmL^*_i)$ and apply
Lemma \ref{lemmadistance} to obtain that
%
%
\begin{eqnarray}
\label{eqderivlowbound2}\qquad
\ddtp(e_{l_p}(\bx,
\rmL_1(t),\ldots,\rmL_K(t)))\bigg|_{t=0} &=&
\lim_{t\rightarrow0}\frac{\dist(\bx, \rmL_j(t))^p - \dist(\bx
,\rmL_j(0))^p}{t^p}\nonumber\\[-8pt]\\[-8pt]
&\geq&-\|\bx\| \lim_{t\rightarrow
0}\frac{\dG(\rmL_j(t),\rmL_j(0))^p}{t^p}.\nonumber
\end{eqnarray}
We also note that for all $0 \leq t \leq t_0$,
%
%
\begin{equation}\label{eqderivlowbound3}
\frac{\dG(\rmL_j(t),\rmL_j(0))^p}{t^p}
\leq
\frac{\dG(\rmL_1(t),\rmL_1(0))^p}{t^p} = 1.
\end{equation}
Indeed, if $t=t_0$, the inequality in (\ref{eqderivlowbound3})
follows from (\ref{eqassume1}) and the equality follows from (\ref
{eqcondparam}).
Moreover, both of them extend to $0 \leq t < t_0$ by the underlying
property of arc length parametrization.
Equation (\ref{eqderivlowbound}) thus follows from (\ref
{eqderivlowbound2}) and (\ref{eqderivlowbound3}).

Combining (\ref{eqderivlowbound}) with Hoeffding's inequality, we
obtain that
%
%
\begin{equation}\label{eqXX0}\quad
\ddtp(e_{l_p}(\rmX_0,
\rmL_1(t),\ldots,\rmL_K(t)))\bigg|_{t=0}\geq-\sum
_{\bx\in\rmX_0}\|\bx\|\geq
-\alpha_0 N \qquad\mbox{w.o.p.}
\end{equation}

We similarly derive an equation analogous to (\ref{eqXX0}) when
replacing $\rmX_0$ with~$\rmX_1$
by applying some arguments of the proof of Lemma \ref{lemmaelldistest}
and Hoeffding's inequality as follows:
%
%
\begin{eqnarray}\label{eqXX1}
\ddtp(e_{l_p}(\rmX_1,
\rmL_1(t),\ldots,\rmL_K(t)))\bigg|_{t=0}&=&\ddt
(e_{l_1}(\rmX_1,
\rmL_1(t)))\bigg|_{t=0}\nonumber\\[-8pt]\\[-8pt]
&\geq& \tau_0 \alpha_1
N \qquad\mbox{w.o.p.}\nonumber
\end{eqnarray}

At last, combining (\ref{eqXX0}), (\ref{eqXX1}) and (\ref{eqcondalpha2}),
we obtain that there exists $\gamma_1' \equiv\gamma_1'(D, d,
K,p,\alpha_0,\alpha_1)$ such that w.o.p.
\[
\ddtp\bigl(e_{l_p}\bigl(\rmX_0\cup\rmX_1,
\rmL_1(t),\ldots,\rmL_K(t)\bigr)\bigr)\bigg|_{t=0}\geq
(\tau_0 \alpha_1 - \alpha_0) N > \gamma_1' N.
\]
Using the arguments of the proof of
\cite{lprecoverypart111}, equation (35), we conclude
that there exists a constant $\gamma_1 \equiv\gamma_1
(D,d,K,p,\alpha_0,\alpha_1,\min_{2\leq i\leq K}\dist(\rmL_1^*,\rmL
_i^*),\break\mu_0,\mu_1)>0$ such that (\ref{eqlocal7}) holds.

\subsection{\texorpdfstring{Conclusion of Theorem \protect\ref{thmhlm2}}{Conclusion of Theorem 1.1}}
\label{sechlm2global}
In order to conclude the theorem, it is enough to prove that $\{\rmL
^*_1,\ldots,\rmL^*_K\}$ is the unique global minimizer
w.o.p. of $e_{l_p}(\rmX_0\cup\rmX_1, \rmL_1,\ldots,\rmL_K)$ in
the set
%
%
\begin{equation}
\label{eqpunctureregion}
\mathrm{GP}(D,d,\gamma_1)
:=\GDd^K\Bigm\backslash
\bigcup_{i_1,i_2,\ldots,i_K\in\calP_K}\ballG((\rmL^*_{i_1},\ldots
,\rmL^*_{i_K}),\gamma_1).\vadjust{\goodbreak}
\end{equation}
%

Combining Proposition \ref{propnonneighbor}, the fact that $d_0>\gamma_1$
[which follows from the definition of $d_0$ in (\ref{eqdefd0})],
Hoeffding's inequality
and (\ref{eqcondalpha2}), we obtain that there exists \mbox{$\gamma_2
\equiv\gamma_2(D,d,K,p,\alpha_0,\min_{1\leq i\leq K}\alpha
_i,\min_{1\leq i\neq j \leq K}\dist(\rmL_i^*,\rmL_j^*),\mu_0,\mu
_1)>0$} such that for any fixed
$(\hat{\rmL}_1,\ldots,\hat{\rmL}_K) \in\mathrm{GP}(D,d,\gamma_1)$,
%
%
\begin{equation}
\label{eqnoneighbor}
e_{l_p}(\rmX,
\hat{\rmL}_1,\ldots,\hat{\rmL}_K)-e_{l_p}(\rmX,
\rmL^*_1,\ldots,\rmL^*_K)>\gamma_2 N \qquad\mbox{w.o.p.}
\end{equation}
Following the proof of \cite{lprecoverypart111}, Theorem 1.1 [i.e.,
covering $\mathrm{GP}(D,d,\gamma_1)$ by balls], we easily
extend (\ref{eqnoneighbor})
w.o.p. for all $K$ subspaces in the set $\mathrm{GP}(D,d,\gamma_1)$
(instead of fixed ones) and thus conclude the theorem.

\section{\texorpdfstring{Proof of Theorem \protect\ref{thmnoisyhlm2} and a counterexample to asymptotic recovery}
{Proof of Theorem 1.2 and a counterexample to asymptotic recovery}}
\label{secnoisyhlm2}

\subsection{\texorpdfstring{Proof of Theorem \protect\ref{thmnoisyhlm2}}{Proof of Theorem 1.2}}
\label{secproofnoisyhlm2}

Following the argument of \cite{lprecoverypart111}, Section 3.5.1, we
reduce the verification of Theorem \ref{thmnoisyhlm2} to proving that
there exists a constant $\gamma_3>0$ such that if for all permutations
$i_1,\ldots,i_K\in\calP_K$, $\hat{\rmL}_1,\ldots, \hat{\rmL}_K\in\GDd$
satisfy that $\dGK((\rmL^*_{i_1}, \ldots,\rmL^*_{i_K})$,
$(\hat{\rmL}_1,\ldots, \hat{\rmL}_K))>f$, then
%
%
\begin{equation}\label{eqnoise2}
\shrtexp_{\mu}(e_{l_p}(\bx,
\hat{\rmL}_1,\ldots,
\hat{\rmL}_K))>\shrtexp_{\mu}(e_{l_p}(\bx,\rmL^*_1,\ldots
,\rmL^*_{K}))+\gamma_3+2\eps^p.
\end{equation}

In view of Proposition \ref{propnonneighbor}, in order to
conclude (\ref{eqnoise2}), it is sufficient to verify
that
%
%
\begin{equation}\label{eqboundnoise21} \Bigl(\tau_0 \min_{1\leq
j\leq
K}\alpha_j-\alpha_0\Bigr)f^p>\gamma_3+2\eps^p
\end{equation}
and
%
%
\begin{equation}\label{eqboundnoise22} \tau_0 \min_{1\leq j\leq
K}\alpha_j
\min_{1\leq i,j\leq
K}\dG^p(\rmL^*_i,\rmL^*_j)/2^p-\alpha_0>\gamma_3+2\eps^p.
\end{equation}
Setting $\gamma_3=\eps^p/2$, (\ref{eqboundnoise21}) follows from
(\ref{eqfdefine}) and (\ref{eqboundnoise22}) follows from
(\ref{eqepsbound}).

\subsubsection{\texorpdfstring{Remark on the size of $\eps$}{Remark on the size of epsilon}}\label{secsizeeps}

If
%
%
\begin{equation}\label{eqnoisebound}
\eps>\pi\sqrt{d} 3^{-{1/p}}\Bigl(\tau_0 \min_{1\leq j\leq
K}\alpha_j-\alpha_0\Bigr)^{1/p}/2,%
\end{equation}
then $f>\pi\sqrt{d}/2$, so that there is no restriction on the
minimizer of (\ref{eqdeferrorksub}) in $\GDd^K$.
It thus makes sense to further restrict $\eps$ to be at least lower
than the right-hand side of (\ref{eqnoisebound}).


\subsection{A counterexample to exact asymptotic recovery with noise}\label{secfapproach0}

\begin{figure}
\begin{tabular}{@{}cc@{}}

\includegraphics{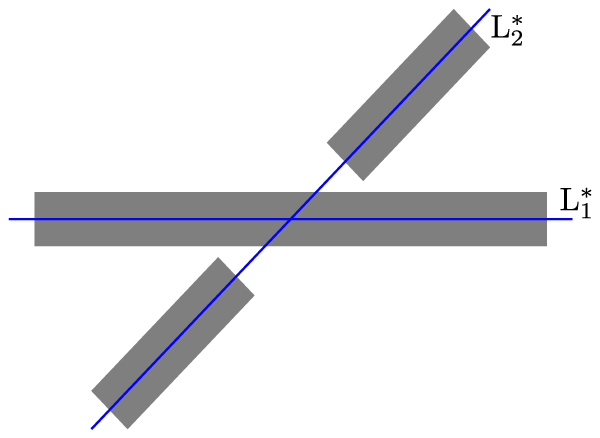}
 & \includegraphics{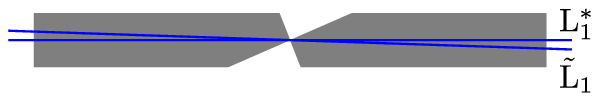}\\
(a) & (b)
\end{tabular}
\vspace*{-3pt}
\caption{A counterexample showing that exact recovery with noise is
impossible even asymptotically. \textup{(a)} Gray regions of uniform
distributions around the two underlying lines. \textup{(b)} The gray
region is the intersection of $\rmY_1$ with the uniform distribution
region around~$\rmL^*_1$. The best $l_p$ line in $\rmY_1$ is
$\tilde{\rmL}_1$.}\label{figsubfigureExample}
\end{figure}

One may ask if it is possible in the noisy setting ($\eps>0$)
to recover the underlying subspaces as the number of sampled points,
$N$, approaches infinity.
The answer to this question is positive when $K=1$
(see, e.g., \cite{Anderson84}, Section
11.6,~\cite{Shawe-taylor05}) or $d=0$ (see
\cite{Pollard82KmeansCLT}).
However, it is often negative when $d>1$ and
\mbox{$K>1$}, as we demonstrate in Figure~\ref{figsubfigureExample}(a) and
%
%
explain below.
In this example, $D=2$, $K=2$, $d=1$, $\alpha_0=0$ and the two
underlying distributions $\mu_1$ and $\mu_2$
(corresponding to the two underlying lines\vadjust{\goodbreak} $\rmL^*_1$ and $\rmL^*_2$)
are uniformly distributed in the two gray regions demonstrated in this figure
(the region around $\rmL_1^*$ is a rectangle and the region around
$\rmL_2^*$ is a union of two disjoint rectangles).

In order to verify that this is indeed a counterexample, we use a
Voronoi-type region, which allows us to reduce approximation by
multiple subspaces to approximation by a single subspace on it.
Such regions $\{\rmY_i\}_{i=1}^{K}$, which are frequently used in
Section \ref{secphasehlm2}, are obtained by a Voronoi diagram
(restricted to the unit ball) of given $d$-subspaces
$\{\rmL_i\}_{i=1}^{K} \subseteq\GDd$ as follows:
%
%
\begin{eqnarray}\label{eqdefrmy1}
&&\rmY_i(\rmL_1,\ldots,\rmL_K) \nonumber\\[-8pt]\\[-8pt]
&&\qquad=
\{\bx\in\ball(\bnull,1)\dvtx\dist(\bx,\rmL_i) < \dist(\bx,\rmL
_j) \mbox{ }
\forall j\dvtx1\leq j\neq i\leq K\}.\nonumber
\end{eqnarray}
These regions are useful to us due to the following elementary
proposition, whose trivial proof is described in the
\hyperref[app]{Appendix}.
\begin{proposition}\label{proprmYibest}
If ${\rmL}'_1,\ldots,{\rmL}'_K \in\GDd$, $\nu$ is a probability
measure on $\reals^D$ and
\[
({\rmL}'_1,\ldots,{\rmL}'_K)=\mathop{\argmin}_{({\rmL}_1,\ldots,{\rmL}_K)
\in\GDd^K}
\shrtexp_{\nu}(e_{l_p}(\bx,
{\rmL}_1,\ldots,{\rmL}_K)),
\]
then
%
%
\begin{equation}\label{eqrmYibest}
\rmL'_1=\mathop{\argmin}_{\rmL_1\in\GDd}\shrtexp_{\nu}\bigl(e_{l
_p}(\bx,
\rmL_1)I\bigl(\bx\in\rmY_1({\rmL}'_1,{\rmL}'_2,\ldots,{\rmL
}'_K)\bigr)\bigr).
\end{equation}
\end{proposition}

We claim that for any fixed $p>0$, the distance between $\{\rmL^*_1,
\rmL^*_2\}$ and the global minimizer of (\ref{eqdeferrorksub}) in the
setting of this example is bounded from below w.o.p. by a positive
constant independent of the sample size, $N$, for sufficiently large $N$.
Equivalently, we claim that the distance between $\{\rmL^*_1, \rmL
^*_2\}$ and the global minimizer of $\shrtexp_{\mu_\eps} (\dist
^p(\bx,\bigcup_{i=1}^K \rmL_i))$ is positive, where $\mu_\eps$ is the
underlying\vadjust{\goodbreak} mixture distribution for this example.
In view of Proposition \ref{proprmYibest}, we only need to show a
positive distance between $\rmL^*_1$ and the minimizer of $\shrtexp
_{\mu_\eps}(e_{l_p}(\bx,
\rmL) I(\bx\in\rmY_1))$, where $\rmY_1=\rmY_1(\rmL_1^*,\rmL
_2^*)$. We refer to this minimizer as the best $l_p$ line for $\rmY
_1$ and denote it by $\tilde{\rmL}_1$ (while arbitrarily fixing $p$).
We note that for any $p>0$, the integral of $l_p$ distances of
points in the part of $\rmY_1$ above $\rmL_1^*$ from the line $\rmL
_1^*$ is smaller than the similar integral in the bottom part.
Therefore, $\tilde{\rmL}_1$ is different than $\rmL^*_1$ and the
respective orientation of the two lines is demonstrated in
Figure \ref{figsubfigureExample}(b). The claim is thus
concluded.


\section{\texorpdfstring{Proof of Theorem \protect\ref{thmphasehlm2}}{Proof of Theorem 1.3}}
\label{secphasehlm2}

\subsection{Preliminaries}\label{secphasehlm2preliminary}
\subsubsection{Notation}

We designate the projection from $\mathbb{R}^D$ onto its
subspace~$\rmL$ by $P_{\rmL}$ and the corresponding orthogonal projection by
$P^\perp_{\rmL}$. We define
%
%
\begin{equation}
\label{eqdefBp1}
\mathbf{D}_{\rmL,\bx,p}=P_{\rmL}(\bx)P^\perp_{\rmL}
(\bx)^{T} \dist(\bx,\rmL)^{(p-2)}.
\end{equation}

We frequently use the Voronoi-type regions $\{\rmY_i\}_{i=1}^{K}$
defined in (\ref{eqdefrmy1}) with respect to the subspaces
$\{\rmL^*_i\}_{i=1}^{K}$ and possibly two additional arbitrary
subspaces denoted by $\hrmlstar\in\GDd$ and $\trml\in\GDd$. We
will use the following short notation for $1\leq i\leq K$:
%
%
\begin{equation}
\label{eqnotationyhat}
\hat{\rmY}_i=\rmY_i(\rmL^*_1,\hrmlstar,\rmL^*_3,\ldots,\rmL
^*_K), \qquad\tilde{\rmY}_i=\rmY_i(\rmL^*_1,\trml,\rmL
^*_3,\ldots,\rmL^*_K)
\end{equation}
and
\begin{equation}
{\rmY}_i=\rmY_i(\rmL^*_1,\rmL^*_2,\rmL
^*_3,\ldots,\rmL^*_K).
\end{equation}

We denote by $\bar{\rmY}_i$ the closure of $\rmY_i$, that is,
%
%
\begin{eqnarray}\label{eqdefbarrmy1}
\bar{\rmY}_i =
\{\bx\in\ball(\bnull,1)\dvtx\dist(\bx,\rmL^*_i) \leq\dist(\bx
,\rmL^*_j) \mbox{ }
\forall j\dvtx1\leq j\neq i\leq K\}.
\end{eqnarray}
Similarly, the closure of $\hat{\rmY}_i$ is denoted by $\bar{\hat
{\rmY}}_i$.

Let $\calL_k$ denote the $k$th-dimensional Lebesgue measure.
We denote $d^*=d \wedge(D-d)$ and let $\theta_{d^*}(\rmL^*_i,\rmL
^*_j)$ be\vspace*{1pt}
the $d^*$th largest principal angle between the $d$-subspaces $\rmL
^*_i$ and $\rmL^*_j$.
Our analysis uses the distribution $\mu\equiv\alpha_0 \mu_0 + \sum
_{i=1}^K\alpha_i\mu_i$,\vspace*{1pt} even though the underlying
distribution of our model is $\mu_\eps$.
For~$\rmL$, $\rmL^* \in\GDd$, we define the ``orthogonal
subtraction'' $\ominus$ as follows:
\[
{\rmL^*}\ominus{\rmL}=\rmL^*\cap(\rmL\cap\rmL^*)^\perp.
\]

\subsubsection{Auxiliary lemmata}
Using the notation above, we formulate two lemmata,
which will be used throughout this proof.
The proof of Lemma~\ref{lemmacombcond3} is identical to that of
\cite{lprecoverypart111}, Proposition 2.2
(while replacing sums by expectations), whereas
Lemma \ref{lemmaintersection} is proved in the \hyperref[app]{Appendix}.
\begin{lemma}
\label{lemmacombcond3} For any $\rmL^* \in\GDd$ and distribution
$\mu$, a necessary condition for $\rmL^*$ to be a local minimum of
$\shrtexp_\mu(l_p(\bx,\rmL))$ is
%
%
\begin{equation}\label{eqnecessary}
\shrtexp_{\mu}(\mathbf{D}_{\rmL^*,\bx,p}) = \bnull.\vadjust{\goodbreak}
\end{equation}
\end{lemma}

The next lemma quantifies the sensitivity of the region $\rmY_j$,
where $1 \leq j \leq K$,
to perturbations in the subspace $\rmL_i$, where $1 \leq i \neq j
\leq K$. WLOG we formulate it with $j=1$ and $i=2$
[note that we use the short notation of~(\ref{eqnotationyhat})].
\begin{lemma}\label{lemmaintersection}
If $\hat{\rmL}_2, \rmL^*_1, \rmL^*_2,\ldots, \rmL^*_K$ are
subspaces in $\GDd$ such that $\hat{\rmL}_2\neq\rmL^*_2$,
%
%
\begin{equation}\label{eqcondition0}
\min_{j\neq2}(\theta_{d^*}(\hat{\rmL}_2,\rmL
^*_j))>0,\qquad 
\min_{1\leq i\neq j\leq K}(\theta_{d^*}(\rmL^*_i,\rmL^*_j))>0
\end{equation}
and
\begin{equation}
\label{eqcondition1}
\theta_{d^*}(\hat{\rmL}_2,\rmL^*_1) \vee\theta_{d^*}(\rmL
^*_2,\rmL^*_1)
\leq\min_{3\leq i\leq K}\theta_{d^*}(\rmL^*_i,\rmL^*_1),
\end{equation}
then
%
%
\begin{equation}\label{eqintersection}
\calL_D\bigl((\hat{\rmY}_1\setminus\rmY_1)\cup
({\rmY}_1\setminus\hat{\rmY}_1)\bigr)>0.
\end{equation}
%
\end{lemma}

\subsection{A special case}
\label{secmotivatingexample} 
\begin{figure}[b]

\includegraphics{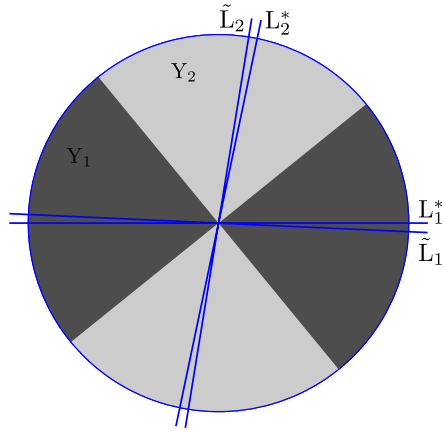}

\caption{Illustrative proof of Theorem \protect\ref{thmphasehlm2} in the
special case where $p=2$, $d=1$, $D=2$ and $K=2$.}\label{figpart2}
\end{figure}

The proof of Theorem \ref{thmphasehlm2} is
rather involved. In order to develop a simple intuition, we provide an
elementary proof of the very special case where $d=1$, $p=2$ and $K=2$.
For simplicity we also assume that $D=2$, though our argument easily
extends to $D > 2$. Figure \ref{figpart2} shows the two underlying
%
%
lines $\rmL^*_1$ and $\rmL^*_2$ and their corresponding regions
$\rmY_1$ and~$\rmY_2$. We note that the best $l_2$ lines [in
$\mathrm{G}(D,1)$] for $\mu_0$ restricted to $\rmY_1$ and~$\rmY_2$ are
the central axes of those regions. Since $\alpha_0>0$, the best $l_2$
lines [in~$\mathrm{G}(D,1)$] for $\mu$ restricted to $\rmY_1$ and
$\rmY_2$ (denoted by $\tilde{\rmL}_1$ and $\tilde {\rmL}_2$, resp.)
must reside between the best $l_2$ lines for $\mu_0$ restricted to
$\rmY_1$ and $\rmY_2$ and $\rmL^*_1$ and $\rmL^*_2$, respectively.
In\vadjust{\goodbreak}
particular,\vspace*{1pt} they are different from $\rmL^*_1$ and
$\rmL^*_2$ as demonstrated in the figure. Therefore,
$\shrtexp_{{\mu}}(e_{l_2}(\bx, \rmL^*_1,\rmL^*_2))
> \shrtexp_{{\mu}}(e_{l_2}(\bx, \tilde{\rmL}_1,\tilde{\rmL}_2))$.
This implies that w.o.p. $e_{l_2}(\rmX, \rmL^*_1,\rmL^*_2) >
e_{l_2}(\rmX, \tilde{\rmL}_1,\tilde{\rmL}_2)$.

\vspace*{-1pt}\subsection{\texorpdfstring{Reduction of the statement of Theorem \protect\ref{thmphasehlm2}
        to simpler formulations}{Reduction of the statement of Theorem 1.3 to simpler formulations}}\vspace*{-1pt}
\label{secphasehlm2reduce}

\subsubsection{Reduction \textup{I}: Using the Voronoi-type regions $\{{\mathrm{Y}}_i\}_{i=1}^K$}

$\!\!\!$We will show here that the following equation implies Theorem \ref
{thmphasehlm2}:
%
%
\begin{equation}
\label{eqconjecture}\qquad
\gamma_{D,d}^K\bigl(
\{\rmL^*_i\}_{i=1}^{K}\subset\GDd\dvtx
\shrtexp_{\mu_0}\bigl(I(\bx\in\rmY_j) \mathbf{D}_{\rmL^*_j,\bx
,p}\bigr)=\bnull\ \forall1\leq j \leq K\bigr)=0.
\end{equation}

First, we apply the argument of \cite{lprecoverypart111},
Section 3.6.1
(which requires the assumption specified in Section~\ref{subsecprecise}
that the
first moments of $\{\|\nu_{i,\eps}\|\}_{i=1}^K$ are smaller than~$\eps$)
to obtain that Theorem
\ref{thmphasehlm2} follows by the equation
%
%
\begin{eqnarray}\label{eqprob2equal0}
&&\gamma_{D,d}^{K}\Bigl(\{\rmL^*_i\}_{i=1}^{K}\subset
\GDd\dvtx
(\rmL^*_1,\ldots,\rmL^*_K) \nonumber\\[-8pt]\\[-8pt]
&&\hphantom{\gamma_{D,d}^{K}\Bigl(}
=\mathop{\argmin}_{({\rmL}_1,\ldots,{\rmL}_K)}
\shrtexp_{\mu}(e_{l_p}(\bx,
{\rmL}_1,\ldots,{\rmL}_K))\Bigr)=0.\nonumber
\end{eqnarray}

Next, applying Proposition \ref{proprmYibest},
we conclude that (\ref{eqprob2equal0}) is a direct consequence of the equation:
%
%
\begin{eqnarray}\label{eqrmL1min}
&&\gamma_{D,d}^K\Bigl(\{\rmL^*_i\}_{i=1}^{K}\subset\GDd\dvtx
\rmL^*_j=\mathop{\argmin}_{\rmL\in\GDd}\shrtexp_{\mu}\bigl(e_{l
_p}(\bx,
\rmL)I(\bx\in\rmY_j)\bigr)\nonumber\\[-8pt]\\[-8pt]
&&\qquad\hspace*{210.3pt} \forall1\leq j \leq
K\Bigr)=0.\nonumber
\end{eqnarray}
Furthermore, applying Lemma \ref{lemmacombcond3} with $\mu= \mu
|_{\rmY_j}$, we obtain that (\ref{eqrmL1min})
follows by the equation
%
%
\begin{equation}\label{eqsimconjecture}\qquad
\gamma_{D,d}^K\bigl(\{\rmL^*_i\}_{i=1}^{K} \subset\GDd\dvtx
\shrtexp_{\mu}\bigl(I(\bx\in\rmY_j)
\mathbf{D}_{\rmL^*_j,\bx,p}\bigr)=0\mbox{ } \forall1\leq j \leq K
\bigr)=0.
\end{equation}

At last we conclude the desired reduction by noting that (\ref
{eqsimconjecture}) and (\ref{eqconjecture}) are equivalent [indeed, the
only relevant components of the distribution $\mu$ in (\ref
{eqsimconjecture}) are $\mu_0$ and $\mu_j$ and the corresponding
expectation according to
$\mu_j$ is zero].

\subsubsection{Reduction \textup{II}: From $K$ subspaces to a single subspace}

We redu\-ce~(\ref{eqconjecture}) so that its underlying condition
involves a single subspace as follows:
%
%
\begin{eqnarray}
\label{eqconjecture1}
&&\gamma_{D,d}\Bigl(\rmL^*_2\in\GDd\dvtx\min_{1\leq i\neq j \leq K}\theta
_{d^*}(\rmL^*_i,\rmL^*_j)>0,\nonumber\\[-8pt]\\[-8pt]
&&\hphantom{\gamma_{D,d}\Bigl(}
\mathop{\argmin}_{2\leq i\leq K}\theta_{d^*}
(\rmL^*_1,\rmL^*_i)=2,
\shrtexp_{\mu_0}\bigl(I(\bx\in\rmY_1)
\mathbf{D}_{\rmL^*_1,\bx,p}\bigr)=\bnull\Bigr)=0.
\nonumber
\end{eqnarray}
We remark that some of the underlying technical conditions of (\ref
{eqconjecture1})
appear in (\ref{eqcondition0}) and (\ref{eqcondition1}) and will be
better understood later when applying Lemma~\ref{lemmaintersection}.

We verify this reduction as follows. WLOG (\ref{eqconjecture1}) can be
formulated by replacing $\rmL^*_2$ with $\rmL^*_k$, for some $3 \leq
k \leq K$,
while letting\vadjust{\goodbreak} $\argmin_{2\leq i\leq K}\theta_{d^*}(\rmL^*_1,\break\rmL^*_i)=k$.
Combining this observation with elementary properties of distributions,
we have that
\begin{eqnarray*}
&&\gamma_{D,d}^K\bigl(\{\rmL^*_i\}_{i=1}^{K} \subset\GDd\dvtx
\shrtexp_{\mu_0}\bigl(I(\bx\in\rmY_j) \mathbf{D}_{\rmL^*_j,\bx
,p}\bigr)=\bnull\mbox{ } \forall1\leq j \leq K \bigr)\\[-1pt]
&&\qquad\leq\sum_{k=2}^{K}\int_{
\GDd^{K-1}} \gamma_{D,d}\Bigl(\rmL^*_k\dvtx\min_{1\leq i\neq j\leq
K}\theta_{d^*}(\rmL^*_i,\rmL^*_j)>0,\\[-1pt]
&&\hspace*{50pt}
\mathop{\argmin}_{2\leq i\leq K}\theta_{d^*}(\rmL^*_1,\rmL^*_i)=k, \\[-1pt]
&&\hspace*{50pt}
\shrtexp_{\mu_0}\bigl(I(\bx\in\rmY_1) \mathbf{D}_{\rmL^*_1,\bx
,p}\bigr)=\bnull | \{\rmL^*_i\}_{1 \leq i\neq k \leq K} \Bigr)
\di(\gamma_{D,d}^{K-1}( \{\rmL^*_i\}_{1 \leq i\neq k
\leq K}) )\\[-1pt]
&&\qquad\quad{}+\gamma_{D,d}^K\Bigl(\{\rmL^*_i\}_{i=1}^{K} \subset\GDd\dvtx\min
_{1\leq i,j\leq K}\theta_{d^*}(\rmL^*_i,\rmL^*_j)=0\Bigr)=0.
\end{eqnarray*}


\vspace*{-2pt}\subsection{Concluding the cases $d=1$ and $d=D-1$}\label{secphasehlm2case1}

We assume first that $d=1$. We conclude the theorem in this case by
proving (\ref{eqconjecture1}) and then extend the analysis to the case $d=D-1$.

\vspace*{-2pt}\subsubsection{\texorpdfstring{Reduction of (\protect\ref{eqconjecture1}) using additional condition
    on the Grassmannian}{Reduction of (52) using additional condition on the Grassmannian}}
\label{secreducebycond1}
We fix $\mathbf{v}_1$ to be one of the two unit vectors spanning $\rmL
^*_1$ and denote by $\mathbf{u}_1$ the unit vector spanning $(\rmL
^*_1+\rmL^*_2)\cap\rmL^{*\perp}_1$ having orientation such that for
any point $\bx\in\rmL^*_2\dvtx(\bx^T\mathbf{u}_1) (\bx^T\mathbf
{v}_1)\geq0$.
We will prove that (\ref{eqconjecture1}) follows from the following
equation, which introduces a restriction on the Grassmannian:
%
%
\begin{eqnarray}\label{eqconjecture2}
&&\gamma_{D,d}\Bigl(\rmL^*_2 \in\GDd\dvtx\min_{1\leq i\neq j\leq K}
\theta_{d^*}(\rmL^*_i,\rmL^*_j)>0,\nonumber\\[-1pt]
&&\hspace*{27.1pt}\mathop{\argmin}_{2\leq i\leq K}\theta_{d^*}(\rmL^*_1,\rmL^*_i)=2,
\\[-1pt]
&&\hspace*{27.1pt}\shrtexp_{\mu_0}\bigl(I(\bx\in\rmY_1) \mathbf{D}_{\rmL^*_1,\bx,p}\bigr)
=\bnull{|} (\rmL^*_1+\rmL^*_2)\cap\rmL^{*\perp}_1=\Sp
(\mathbf{u}_1)\Bigr)=0.\nonumber
\end{eqnarray}

%
\begin{figure}[b]

\includegraphics{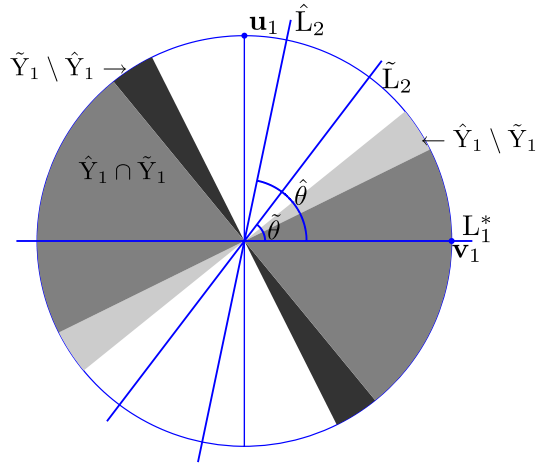}

\caption{The regions $\hat{\rmY}_1$ and $\tilde{\rmY}_1$ and the
relation to $\hat{\theta}$ and $\tilde{\theta}$ when $d=1$ and
$K=2$.}\label{figfigure1}
\end{figure}

We define the following subset of the sphere $S^{D-1}\dvtx\Omega_0 = \{
\bx\in S^{D-1}\dvtx\bx\perp\mathbf{v}\}$,
and a distribution $\omega$ on $\Omega_0$ such that for any $\sA
\subseteq\Omega_0\dvtx\omega(\sA)=\gamma_{D,d}(\rmL^*_2\in\GDd
\dvtx(\rmL^*_1+\rmL^*_2)\cap\rmL^{*\perp}_1\in
\Sp(\sA))$.
Using this notation, (\ref{eqconjecture2}) implies (\ref
{eqconjecture1}) as follows:
\begin{eqnarray*}
\hspace*{-4pt}&&
\gamma_{D,d}\Bigl(\rmL^*_2\in\GDd\dvtx\min_{1\leq i\neq j \leq
K}\theta_{d^*}(\rmL^*_i,\rmL^*_j)>0, \mathop{\argmin}_{2\leq i\leq
K}\theta_{d^*}(\rmL^*_1,\rmL^*_i)=2, \\[-1pt]
\hspace*{-4pt}&&\hspace*{193.5pt}
\shrtexp_{\mu_0}\bigl(I(\bx\in\rmY_1) \mathbf{D}_{\rmL^*_1,\bx
,p}\bigr)=\bnull\Bigr)\\[-1pt]
\hspace*{-4pt}&&\qquad=\int_{\Omega_0}\gamma_{D,d}\Bigl(\rmL^*_2\dvtx\min_{1\leq i\neq
j\leq K}\theta_{d^*}(\rmL^*_i,\rmL^*_j)>0, \mathop{\argmin}_{2\leq i\leq
K}\theta_{d^*}(\rmL^*_1,\rmL^*_i)=2,\\[-1pt]
\hspace*{-4pt}&&\qquad\hspace*{40pt} \shrtexp_{\mu_0}
\bigl(I(\bx\in\rmY_1) \mathbf{D}_{\rmL^*_1,\bx,p}\bigr)=\bnull
{|}(\rmL^*_1+\rmL^*_2)\cap\rmL^{*\perp}_1=
\Sp(\mathbf{u}_1)\Bigr)\,\di(\omega(\mathbf{u}_1))\\[-1pt]
\hspace*{-4pt}&&\qquad=0.\vadjust{\goodbreak}
\end{eqnarray*}

\subsubsection{\texorpdfstring{Proof of (\protect\ref{eqconjecture2})}{Proof of (53)}}

We will show that at most one element satisfies the underlying
condition of (\ref{eqconjecture2}) (i.e., it is a member of the set
for which~$\gamma_{D,d}$ is evaluated).
Assume, on the contrary, that there are
two\vspace*{2pt} subspaces $\hrmlstar$ and~$\trml$ satisfying this condition with
corresponding angles
$\hat{\theta}=\theta_{d^*}(\rmL_1^*,\hrmlstar)$ and $\tilde
{\theta}=\theta_{d^*}(\rmL_1^*,\trml)$ in $[0,\pi/2]$, where WLOG
$\hat{\theta}>\tilde{\theta}$.
Using the notation of~(\ref{eqnotationyhat}), we have that
%
%
\begin{eqnarray}\label{eqhattildermY}
&&\shrtexp_{\mu_0}\bigl(I(\bx\in\tilde{\rmY}_1\setminus\hat
{\rmY}_1) \mathbf{D}_{\rmL^*_1,\bx,p}\bigr)-\shrtexp_{\mu
_0}\bigl(I(\bx\in\hat{\rmY}_1\setminus\tilde{\rmY}_1) \mathbf
{D}_{\rmL^*_1,\bx,p}\bigr)\nonumber\\
&&\qquad= 2 \cdot\bigl(\shrtexp_{\mu
_0}\bigl(I(\bx\in\tilde{\rmY}_1) \mathbf{D}_{\rmL^*_1,\bx,p}\bigr) -
\shrtexp_{\mu_0}\bigl(I(\bx\in\hat{\rmY}_1) \mathbf{D}_{\rmL
^*_1,\bx,p}\bigr)\bigr)\\
&&\qquad=\bnull-\bnull= \bnull.\nonumber
\end{eqnarray}
Consequently,
%
%
\begin{equation}\label{eqhattildermYbvu}
\shrtexp_{\mu_0}\bigl(I(\bx\,{\in}\,\tilde{\rmY}_1\,{\setminus}\,\hat{\rmY
}_1) \mathbf{v}_1^T\mathbf{D}_{\rmL^*_1,\bx,p}\mathbf{u}_1
\bigr)\,{-}\,\shrtexp_{\mu_0}\bigl(I(\bx\,{\in}\,\hat{\rmY}_1\,{\setminus}\,\tilde
{\rmY}_1) \mathbf{v}_1^T\mathbf{D}_{\rmL^*_1,\bx,p}\mathbf
{u}_1\bigr)\,{=}\,\bnull.\hspace*{-32pt}
\end{equation}

Defining
\[
\theta_{\mathbf{u}_1,\mathbf{v}_1}(\bx)=\mathrm{arctan}\frac
{\mathbf{u}_1\cdot\bx}{\mathbf{v}_1\cdot\bx}
\]
and
\[
\rmY_{1,\hat{2}}=\Bigl\{\bx\in\ball(\mathbf{0},1)\dvtx\dist(\bx,\rmL
^*_1)<\min_{3 \leq i\leq K} \dist(\bx,\rmL^*_i) \Bigr\},
\]
we express the regions $\hat{\rmY}_1$ and $\tilde{\rmY}_1$ as follows:
%
%
\begin{eqnarray}\label{eqfig11}
\hat{\rmY}_1&=&\rmY_{1,\hat{2}}\cap\{\bx\in\ball
(\bnull,1)\dvtx
\hat{\theta}/2-\pi/2<\theta_{\mathbf{u}_1,\mathbf{v}_1}(\bx
)<\hat{\theta}/2\},
\\
\label{eqfig12}
%
\tilde{\rmY}_1&=&\rmY_{1,\hat{2}}\cap\{\bx\in\ball
(\bnull
,1)\dvtx\tilde{\theta}/2-\pi/2<\theta_{\mathbf{u}_1,\mathbf
{v}_1}(\bx)<\tilde{\theta}/2\}.
\end{eqnarray}
Figure \ref{figfigure1} clarifies (\ref{eqfig11}) and (\ref
{eqfig12}) in the special case where $d=1$\vadjust{\goodbreak} and
$K=2$.

Combining (\ref{eqfig11}) and (\ref{eqfig12}) with the definition of
$\mathbf{D}_{\rmL,\bx,p}$ in (\ref{eqdefBp1}), we
obtain that
%
%
\begin{equation}\label{eqhatrmY}\quad
\hat{\rmY}_1\setminus\tilde{\rmY}_1\subset\bigl\{\bx\in\ball(\bnull
,1)\dvtx\mathbf{v}_1^T\bx\bx^T\mathbf{u}_1 \equiv\dist(\bx,\rmL
^*_1)^{(2-p)}\mathbf{v}_1^T\mathbf{D}_{\rmL^*_1,\bx,p}\mathbf
{u}_1>0\bigr\}\hspace*{-20pt}
\end{equation}
and
%
%
\begin{equation}\label{eqtildermY}\quad
\tilde{\rmY}_1\setminus\hat{\rmY}_1\subset\bigl\{\bx\in\ball(\bnull
,1)\dvtx\mathbf{v}_1^T\bx\bx^T\mathbf{u}_1 \equiv\dist(\bx,\rmL
^*_1)^{(2-p)}\mathbf{v}_1^T\mathbf{D}_{\rmL^*_1,\bx,p}\mathbf
{u}_1<0\bigr\}.\hspace*{-20pt}
\end{equation}

It follows from Lemma \ref{lemmaintersection} that $\calL_D ((\tilde
{\rmY}_1\setminus\hat{\rmY}_1)\cup(\hat{\rmY}_1\setminus\tilde
{\rmY}_1))>0$ and, consequently, for any $r>0$, $\calL_D(\ball
(\bnull,r)\cap((\tilde{\rmY}_1\setminus\hat{\rmY}_1)\cup(\hat
{\rmY}_1\setminus\tilde{\rmY}_1)))>0$
(indeed, if $\bx\in\rmY_1$, then $c\cdot\bx\in\rmY_1$ for any
$0<c<1/\|\bx\|$; thus, the distribution in the latter inequality is
just a scaling by $r^D$ of the distribution in the former one). Since
there exists $r>0$ such that the restriction of $\calL_D$ to $\ball
(\bnull,r)$ is
absolutely continuous with respect to $\mu_0$,
we also have that
$\mu_0(\ball(\bnull,r)\cap((\tilde{\rmY}_1\setminus\hat{\rmY
}_1)\cup(\hat{\rmY}_1\setminus\tilde{\rmY}_1)))>0$.
However, this contradicts (\ref{eqhattildermYbvu}), (\ref{eqhatrmY})
and (\ref{eqtildermY}), that is, it
proves (\ref{eqconjecture2}) and therefore the theorem in the current
special case.\looseness=1

\subsubsection{The case $d=D-1$}
We note that the proof of the above case ($d=1$) can be adapted to the
case where $d=D-1$. This is done by letting $\mathbf{v}_1$ be one of
the two unit vectors spanning $\rmL^*_1\cap(\rmL^*_1\cap\rmL
^*_2)^\perp$ [note that $\dim(\rmL^*_1)=D-1$ and $\dim(\rmL
^*_1\cap\rmL^*_2) =d-2$ so that
$\dim(\rmL^*_1\cap(\rmL^*_1\cap\rmL^*_2)^\perp)=1$] and $\mathbf
{u}_1$ be the unit vector of $(\rmL^*_1+\rmL^*_2)\cap\rmL_1^\perp$
with a similar orientation as in the case where $d=1$.

\subsection{\texorpdfstring{Conclusion: The case where $d \neq1$ and $d \neq D-1$}
{Conclusion: The case where d/=1 and d/=D-1$}}\label{secphasehlm2case2}

\subsubsection{\texorpdfstring{Reduction of (\protect\ref{eqconjecture1}) using
additional condition on the Grassmannian}{Reduction of (52) using additional condition on the Grassmannian}}
\label{secreducebycond2}

The following reduction is analogous to the one of Section \ref
{secreducebycond1}.
Denoting by $B(\mathbb{R}^{D} , \mathbb{R}^{D})$ the space of linear operators
from $\mathbb{R}^{D}$ to itself, we define
\begin{eqnarray*}
\Omega_1&=&\{(P_1,P_2)\in B(\mathbb{R}^{D} , \mathbb{R}^{D})^2\dvtx
\exists \rmL\in\GDd
\mbox{ not orthogonal to } \rmL^*_1,\\
&&\hspace*{23.5pt} \mbox{s.t. } \dim(\rmL^*_1
\ominus\rmL)>1, P_{\rmL^*_1}^TP_{\rmL}P_{\rmL^*_1}=P_1,
P_{\rmL^*_1}^{\perp T}P_{\rmL}P^\perp_{\rmL^*_1}=P_2\}
\end{eqnarray*}
and the distribution $\omega_1$ on $\Omega_1$ as follows: for any set
$\sA\subseteq\Omega_1$,
\[
\omega_1(\sA)=\gamma_{D,d}\bigl(\rmL\in\GDd\dvtx(P_{\rmL
^*_1}^TP_{\rmL}P_{\rmL^*_1},
P_{\rmL^*_1}^{\perp T}P_{\rmL}P^\perp_{\rmL^*_1}) \in\sA\bigr).
\]

Using this notation, we reduce (\ref{eqconjecture1}) as follows:
%
%
\begin{eqnarray}\label{eqconjecture3}
&&\gamma_{D,d}\Bigl(\rmL^*_2 \in\GDd\dvtx\rmL^*_1 \not\perp\rmL
^*_2, \operatorname{dim}(\rmL^*_1\cap\rmL^{*\perp}_2)>1,\nonumber\\
&&\hphantom{\gamma_{D,d}\Bigl(}
\min_{1\leq i\neq j \leq K}\theta_{d^*}(\rmL^*_i,\rmL^*_j)>0,
\mathop{\argmin}_{2\leq i\leq
K}\theta_{d^*}(\rmL^*_1,\rmL^*_i)=2,\nonumber\\[-8pt]\\[-8pt]
&&\hphantom{\gamma_{D,d}\Bigl(}
\shrtexp_{\mu_0}\bigl(I(\bx\in\rmY_1) \mathbf{D}_{\rmL^*_1,\bx
,p}\bigr)=\bnull|\nonumber\\
&&\hphantom{\gamma_{D,d}\Bigl(}
\hspace*{31pt}(P_{\rmL^*_1}^TP_{\rmL^*_2}P_{\rmL^*_1},
P_{\rmL^*_1}^{\perp T}P_{\rmL^*_2}P^\perp_{\rmL
^*_1})=(P_1,P_2)\in\Omega_1
\Bigr)=0.\nonumber
\end{eqnarray}
Indeed,\vspace*{-3pt}
\begin{eqnarray*}
&&\gamma_{D,d}\Bigl(\rmL^*_2\in\GDd\dvtx\min_{1\leq i\neq j \leq
K}\theta_{d^*}(\rmL^*_i,\rmL^*_j)>0, \mathop{\argmin}_{2\leq i\leq
K}\theta_{d^*}(\rmL^*_1,\rmL^*_i)=2,\\[-2pt]
&&\hspace*{193.5pt}
\shrtexp_{\mu_0}\bigl(I(\bx\in\rmY_1) \mathbf{D}_{\rmL^*_1,\bx
,p}\bigr)=\bnull\Bigr)\\[-2pt]
&&\qquad\leq\int_{\Omega_1}\gamma_{D,d}\bigl(\rmL^*_2\dvtx\rmL^*_1 \mbox
{ is not orthogonal to }
\rmL^*_2,\\[-2pt]
&&\qquad\hphantom{\leq\int_{\Omega_1}\gamma_{D,d}\bigl(}
\dim(\rmL^*_1 \ominus\rmL^*_2)>1,
\min_{1\leq i\neq j
\leq K}\theta_{d^*}(\rmL^*_i,\rmL^*_j)>0,\\[-2pt]
&&\qquad\hphantom{\leq\int_{\Omega_1}\gamma_{D,d}\bigl(}
 \mathop{\argmin}_{2\leq i\leq
K}\theta_{d^*}(\rmL^*_1,\rmL^*_i)=2,\\[-2pt]
&&\qquad\hphantom{\leq\int_{\Omega_1}\gamma_{D,d}\bigl(}
\shrtexp_{\mu_0}\bigl(I(\bx\in\rmY_1) \mathbf{D}_{\rmL^*_1,\bx
,p}\bigr)=\bnull|\\[-2pt]
&&\qquad\hphantom{\leq\int_{\Omega_1}\gamma_{D,d}\bigl(}
\hspace*{7pt}
(P_{\rmL^*_1}^TP_{\rmL^*_2}P_{\rmL^*_1},
P_{\rmL^*_1}^{\perp T}P_{\rmL^*_2}P^\perp_{\rmL^*_1})
=(P_1,P_2)\in\Omega_1
\bigr)\,\di(\omega_1(P_1,P_2))\\[-2pt]
&&\qquad\quad{}
+\gamma_{D,d}\bigl(\rmL^*_2\in\GDd\dvtx\dim(\rmL^*_1 \ominus\rmL^*_2)
\leq1\mbox{, or } \rmL^*_2\perp\rmL^*_1\bigr)
=0+0=0.\vspace*{-1pt}
\end{eqnarray*}

\subsubsection{Bulk of the proof}

We prove (\ref{eqconjecture3}) by using the following two lemmata,
which are proved below (Sections \ref{secinfiniteelements}
and \ref{sechattildermL2}).\vspace*{-3pt}
\begin{lemma}\label{lemmainfiniteelements}
If $\dim(\rmL^*_1 \ominus\rmL^*_2)\geq2$ and $\rmL^*_1$ is not
orthogonal to $\rmL^*_2$, then the set\vspace*{-1pt}
\[
\sZ=\{\rmL\in\GDd\dvtx P_{\rmL^*_1}(P_{\rmL
^*_2}-P_{\rmL})P_{\rmL^*_1}=0, P^\perp_{\rmL^*_1}(P_{{\rmL
}^*_2}-P_{\rmL})P^\perp_{\rmL^*_1}=0\}\vspace*{-1pt}
\]
is infinite.\vspace*{-3pt}
\end{lemma}
\begin{lemma}\label{lemmahattildermL2}
$\!\!\!$If $\trml, \hrmlstar\,{\in}\,\GDd$ satisfy $\trml\,{\neq}\,\hrmlstar$,
$\theta_{d^*}(\hrmlstar,\rmL^*_1) \vee
\theta_{d^*}(\rmL^*_2,\rmL^*_1)\leq\min_{3\leq i\leq K}\theta_{d^*}(\rmL
^*_i,\rmL
^*_1)$, $P_{\rmL^*_1}(P_{\hrmlstar}-P_{\trml})P_{\rmL^*_1}=0$ and
$P^{\perp}_{\rmL^*_1}(P_{\hrmlstar}-P_{\trml})P^\perp_{\rmL
^*_1}=0$, then either $\hrmlstar$ or $\trml$ will not satisfy the
condition in (\ref{eqconjecture3}).\vspace*{-3pt}
\end{lemma}

To conclude (\ref{eqconjecture3}), we rewrite it as follows:
$\gamma_{D,d}(A|B)=0$, where $A$ and~$B$
are clear from the context. We note that Lemma \ref{lemmainfiniteelements}
implies that there
are infinitely many subspaces $\rmL^*_2$ in $B$. On the other hand,
Lemma \ref{lemmahattildermL2} implies that there is only one subspace
$\rmL^*_2$ in $A$.
These observations clearly prove~(\ref{eqconjecture3}).
We remark that the idea of this proof is somewhat similar to
that of the previous case where $d = 1$ or $d=D-1$. In this case,
Lemma \ref{lemmainfiniteelements} is analogous to the fact that there
is a degree of freedom in choosing $\rmL_2^*$ in~(\ref
{eqconjecture2}) [since we can choose any $\theta_{d^*}(\rmL_1^*,\rmL
_2^*)<\min_{3\leq i\leq K}\theta_{d^*}(\rmL_1^*,\rmL_i^*)$].
Moreover, Lemma \ref{lemmahattildermL2} is analogous to the fact that
there were not two subspaces $\hrmlstar$ and $\trml$ satisfying the
underlying condition of (\ref{eqconjecture2}).\vspace*{-3pt}

\subsubsection{\texorpdfstring{Proof of Lemma \protect\ref{lemmainfiniteelements}}{Proof of Lemma 4.3}}
\label{secinfiniteelements}
We denote $\tilde{\rmL}_1={\rmL^*_1}\ominus{(\rmL^*_1\cap\rmL
^*_2)}$ and $\tilde{\rmL}_2=\rmL^*_2\ominus(\rmL^*_1\cap\rmL^*_2)$.
The idea of the proof is to construct a one-to-one function
$g\dvtx S^{D-1}\cap\tilde{\rmL}_2\rightarrow\sZ$.
Then, using this function and the fact that $\operatorname{dim}(\tilde
{\rmL
}_2)=\operatorname{dim}(\rmL^*_1)-\operatorname{dim}(\rmL^*_2\cap\rmL
^*_1)\geq2$,
we conclude that $\sZ$, which contains $g(S^{D-1}\cap\tilde{\rmL
}_2)$,\vadjust{\goodbreak} is infinite.

For any $\mathbf{u}_0\in S^{D-1}\cap\tilde{\rmL}_2$, we arbitrarily
fix $\mathbf{v}_0=\mathbf{v}_0(\mathbf{u}_0)$ as one of the two unit
vectors spanning
$ \tilde{\rmL}_1\cap(\tilde{\rmL}_2\ominus\Sp(\mathbf
{u}_0))^\perp$. The vector $\mathbf{v}_0$ exists since
\begin{eqnarray*}
\dim\bigl(\tilde{\rmL}_1\cap\bigl(\tilde{\rmL}_2\ominus\Sp
(\mathbf{u}_0)\bigr)^\perp\bigr) &\geq& \dim(\tilde{\rmL}_1)+
\dim\bigl(\bigl(\tilde{\rmL}_2\ominus\Sp(\mathbf{u}_0)
\bigr)^\perp\bigr)-D\\
&=&d+(D-d+1)-D=1.
\end{eqnarray*}
We define the function $g$ as follows:
\[
g(\mathbf{u}_0)=\Sp\bigl(\mathbf{u}_0-2(\mathbf{v}_0^T\mathbf
{u}_0)\mathbf{v}_0,\rmL^*_2\ominus\Sp(\mathbf{u}_0)\bigr).
\]

We first claim that the image of $g$ is contained in $\sZ$. Indeed, we
note that
%
%
\begin{eqnarray}
\label{eqdiffpg}\quad
P_{g(\mathbf{u}_0)}-P_{\rmL^*_2}&=&\bigl(\mathbf{u}_0-2(\mathbf
{v}_0^T\mathbf{u}_0)\mathbf{v}_0\bigr)^T
\bigl(\mathbf{u}_0-2(\mathbf{v}_0^T\mathbf{u}_0)\mathbf{v}_0\bigr)-\mathbf
{u}_0^T\mathbf{u}_0\nonumber\hspace*{-20pt}\\[-8pt]\\[-8pt]
&=&-2(\mathbf{v}_0^T\mathbf{u}_0)
\bigl(\mathbf{v}_0^T\bigl(\mathbf{u}_0-(\mathbf{v}_0^T\mathbf
{u}_0)\mathbf{v}_0\bigr)+\bigl(\mathbf{u}_0-(\mathbf{v}_0^T\mathbf
{u}_0)\mathbf{v}_0\bigr)^T\mathbf{v}_0\bigr) \hspace*{-20pt}\nonumber.
\end{eqnarray}
Combining (\ref{eqdiffpg}) with the following two facts: $\mathbf
{v}_0\in\rmL^*_1$ and $\mathbf{u}_0-(\mathbf{v}_0^T\mathbf
{u}_0)\mathbf{v}_0\in\rmL^{*\perp}_1$, we obtain that
$g(\mathbf{u}_0)\in\sZ$.

At last, we prove that $g$ is one-to-one and thus conclude the proof.
If, on the contrary, there exist $\mathbf{u}_1$, $\mathbf{u}_2\in
S^{D-1}\cap\tilde{\rmL}_2$ such that $\mathbf{u}_1 \neq\mathbf
{u}_2$ and $g(\mathbf{u}_1)=g(\mathbf{u}_2)$,
then $g(\mathbf{u}_1)=\Sp(g(\mathbf{u}_1),g(\mathbf{u}_2))
\supseteq(\rmL^*_2\ominus\Sp(\mathbf{u}_1)) +
( \rmL^*_2\ominus\Sp(\mathbf{u}_2))\supseteq\rmL^*_2$.
Since $\dim(g(\mathbf{u}_1))=\dim(\rmL^*_2)$, we conclude
that $g(\mathbf{u}_1)=\rmL^*_2$. On the other hand, we claim that for
any $\mathbf{u}_0 \in S^{D-1}\cap\tilde{\rmL}_2\dvtx
g(\mathbf{u}_0)\neq\rmL^*_2$ and thus obtain a~contradiction. Indeed,
since $\mathbf{u}_0\in\tilde{\rmL}_2$, $\mathbf{v}_0\in\tilde
{\rmL}_1$ and $\rmL^*_1$ is not orthogonal to~$\rmL^*_2$, we have
that $\mathbf{v}_0^T\mathbf{u}_0 \neq0$
and, consequently, $\mathbf{u}_0-(\mathbf{v}_0^T\mathbf{u}_0)\mathbf
{v}_0\neq\mathbf{u}_0$.
Applying the latter observation in (\ref{eqdiffpg}), we obtain that
$P_{g(\mathbf{u}_0)}\neq P_{\rmL^*_2}$ and, consequently, $g(\mathbf
{u}_0)\neq\rmL^*_2$.

\subsubsection{\texorpdfstring{Proof of Lemma \protect\ref{lemmahattildermL2}}{Proof of Lemma 4.4}}
\label{sechattildermL2}
We assume, on the contrary, that both $\hrmlstar$ and $\trml$ satisfy
the underlying condition of (\ref{eqconjecture1}) and conclude a
contradiction.

We arbitrarily fix here $\bx\in\hat{\rmY}_1\setminus\tilde{\rmY
}_1$ [using the notation of (\ref{eqnotationyhat})]. We note that
$\dist(\bx,\rmL^*_1)<\dist(\bx,\hrmlstar)$
and $\dist(\bx,\rmL^*_1)<\argmin_{3\leq i\leq K}\dist(\bx,\rmL
^*_i)$. Sin\-ce $\bx\notin\tilde{\rmY}_1$, we have that $\dist(\bx
,\rmL^*_1)>\dist(\bx,\trml)$
and, thus,
%
%
\begin{equation}\label{eqcomparisonrmY1}
\dist(\bx,\trml)<\dist(\bx,\rmL^*_1)<\dist(\bx,\hrmlstar).
\end{equation}
Consequently,
%
%
\begin{equation}\label{eqcomparisonrmY2}
\bx^T(P_{\hrmlstar}-P_{\trml})\bx=\dist(\bx,\trml)^2- \dist(\bx
,\hrmlstar)^2<0.
\end{equation}

We partition $P_{\hrmlstar}-P_{\trml}$ into four parts: $P_{\rmL
^*_1}(P_{\hrmlstar}-P_{\trml})P_{\rmL^*_1}$,
$P^{\perp}_{\rmL^*_1}(P_{\hrmlstar}-P_{\trml})P^\perp_{\rmL
^*_1}$, $P_{\rmL^*_1}(P_{\hrmlstar}-P_{\trml})P^\perp_{\rmL^*_1}$
and $P^{\perp}_{\rmL^*_1}(P_{\hrmlstar}-P_{\trml})P_{\rmL^*_1}$.
The first\vspace*{-1pt} two are zero, and the last two are adjoint to each other; we
thus only consider $P_{\rmL^*_1}(P_{\hrmlstar}-P_{\trml})P^\perp
_{\rmL^*_1}$.
Let its SVD be
%
%
\begin{equation}
\label{eqsvdtotdiffpart}
P_{\rmL^*_1}(P_{\hrmlstar}-P_{\trml})P^\perp_{\rmL^*_1}=\bU
\bolds{\Sigma}\bV=\sum_{i=1}^{d}\sigma_i\mathbf{u}_i\mathbf{v}_i^T.
\end{equation}
We can express the SVD of $P_{\hrmlstar}-P_{\trml}$ using (\ref
{eqsvdtotdiffpart}) and the partition above as follows:\vspace*{-3pt}
%
%
\begin{equation}
\label{eqsvdtotdiff}
P_{\hrmlstar}-P_{\trml}=\sum_{i=1}^{d}\sigma_i(\mathbf{u}_i\mathbf
{v}_i^T+\mathbf{v}_i\mathbf{u}_i^T).
\end{equation}
Combining (\ref{eqcomparisonrmY2}) and (\ref{eqsvdtotdiff}), we
obtain that
%
%
\begin{equation}
\label{eqcomparisonrmY3}
\sum_{i=1}^{n}\sigma_i\mathbf{u}^{T}_{i}\bx\bx^T\mathbf{v}_i=\bx
^T\Biggl(\sum_{i=1}^{n}\sigma_i(\mathbf{u}_i\mathbf{v}^T_i+\mathbf
{v}_i\mathbf{u}^T_i)\Biggr)\bx/2<0.
\end{equation}

We define a function $f\dvtx\reals^{D \times D} \rightarrow\reals$ such
that for any $\bA\in\reals^{D \times D}\dvtx
f(\bA)=\sum_{i=1}^{n}\sigma_i\mathbf{u}^{T}_{i}\bA\mathbf{v}_i$.
Using (\ref{eqcomparisonrmY3}) and the fact that $\{\mathbf{u}_i\}
_{i=1}^{d}\in\rmL^*_1$ and $\{\mathbf{v}_i\}_{i=1}^{d}\in\rmL
^{*\perp}_1$, we deduce that
%
%
\begin{eqnarray}\label{eqcomparisonrmY4}
f(\bD_{\rmL^*_1,\bx,p})&=&\dist(\bx,\rmL^*_1)^{(p-2)} f(P_{\rmL
^*_1}(\bx)P^\perp_{\rmL^*_1}(\bx)^{T})\nonumber\\[-2pt]
&=& \dist(\bx
,\rmL^*_1)^{(p-2)}\sum_{i=1}^{n}\sigma_i\mathbf{u}^{T}_{i}P_{\rmL
^*_1}(\bx)P^\perp_{\rmL^*_1}(\bx)^T\mathbf{v}_i\\[-2pt]
&=&\dist(\bx,\rmL^*_1)^{(p-2)}\sum_{i=1}^{n}\sigma_i\mathbf
{u}^{T}_{i}\bx\bx^T\mathbf{v}_i<0.\nonumber
\end{eqnarray}
Similarly, for any point $\bx\in\tilde{\rmY}_1\setminus\hat{\rmY}_1$,
%
%
\begin{equation}\label{eqcomparisonrmY5}
f(\bD_{\rmL^*_1,\bx,p})>0.
\end{equation}
Combining (\ref{eqhattildermY}), (\ref{eqcomparisonrmY4}), (\ref
{eqcomparisonrmY5}), Lemma \ref{lemmaintersection} and the linearity
of $f$,
we conclude the following contradiction establishing the current lemma:
%
%
\begin{eqnarray}\label{eqhattildermY1}\quad
0&=&f\bigl(\shrtexp_{\mu_0}\bigl(I(\bx\in\tilde{\rmY}_1\setminus\hat
{\rmY}_1)
\mathbf{D}_{\rmL^*_1,\bx,p}\bigr)-\shrtexp_{\mu_0}\bigl(I(\bx\in\hat{\rmY
}_1\setminus\tilde{\rmY}_1) \mathbf{D}_{\rmL^*_1,\bx,p}\bigr)\bigr)\nonumber\\[-1pt]
&=&f\bigl(\shrtexp_{\mu_0}\bigl(I(\bx\in\tilde{\rmY}_1\setminus\hat
{\rmY}_1) \mathbf{D}_{\rmL^*_1,\bx,p}\bigr)\bigr)-
f\bigl(\shrtexp_{\mu_0}\bigl(I(\bx\in\hat{\rmY}_1\setminus\tilde
{\rmY}_1) \mathbf{D}_{\rmL^*_1,\bx,p}\bigr)\bigr)\\[-1pt]
&>&0.\nonumber\vspace*{-3pt}
\end{eqnarray}

\subsubsection{\texorpdfstring{Remark on the sizes of $\delta_0$ and $\kappa_0$}
{Remark on the sizes of delta0 and kappa0}}
\label{secdeltakappa}
The constants $\delta_0$ and $\kappa_0$ depend on other
parameters of the underlying weak HLM model, in particular, the
underlying subspaces $\{\rmL^*_i\}_{i=1}^{K}$.
For example, one can bound both $\kappa_0$ and
$\delta_0$ from below by the following number:
\[
\max_{1 \leq i \leq K}\Bigl(\shrtexp_{\mu}\bigl(e_{l
_p}(\bx,
\rmL_i^*)I(\bx\in\rmY_i)\bigr)-\min_{\rmL\in\GDd}\shrtexp
_{\mu}\bigl(e_{l_p}(\bx,
\rmL)I(\bx\in\rmY_i)\bigr)\Bigr)\big/ (4p).
\]
%
If $p\geq2$, then a simpler lower bound on both $\kappa_0$ and
$\delta_0$ is
\[
\frac{\|{\max_{1 \leq i \leq K} \shrtexp_{\mu}}(\mathbf
{D}_{\rmL^*_1,\bx,p} I (\bx\in\rmY_i))\|_2^2}{pdD2^{p+5}}.\vadjust{\goodbreak}
\]

\section{Discussion}
\label{secconclusion} We studied the effectiveness of $l_p$
minimization for recovering (or nearly recovering) all
underlying $K$ subspaces for i.i.d. samples from two different types of
HLM distributions.
In particular, we \mbox{demonstrated} a~phase transition phenomenon around $p=1$.
%

We discuss here implications, extensions and limitations of this theory
as well as some open directions.
%

\subsection{Obstacles for convex recovery of multiple subspaces}
\label{secnotconvex} There are some recent methods for robust single
subspace recovery by convex optimization (see, e.g.,
\cite{wrightrobustpca09}). Such methods minimize a real-valued convex
function $h$ on a convex set $\bbH$ (e.g., set of matrices), which can
be mapped on $\GDd$. However, such a minimization cannot be done for
multiple subspaces. Indeed, in that case one must minimize a
multivariate function $h\dvtx\bbH^K \rightarrow\reals$ for convex
$\bbH$. Clearly, the function $h$ must be invariant to permutations of
coordinates. Let $g$ be a mapping of $\bbH$ onto $\GDd$. It follows
from the assumption that the minimization of $h$ leads to the
underlying subspaces $\{\rmL_i^*\}_{i=1}^{K}$ and the
permutation-invariance of $h$ that the set of minimizers of~$h$
coincides with all\vspace*{1pt} permutations of
$\hat{\bx}_1,\hat{\bx}_2,\ldots,\hat{\bx}_K$, where $\hat{\bx}_i\in
g^{-1}(\rmL_i^*)$ for all $1\leq i\leq K$. Since $h$ is convex,
$({\sum_{i=1}^{K}\hat{\bx}_i}/{K},\ldots,{\sum_{i=1}^{K}\hat{\bx
}_i}/{K})$
is also a minimizer of $h$. Consequently, ${\sum_{i=1}^{K}\hat{\bx
}_i}/{K}\in
g^{-1}(\rmL_j^*)$ for all $1\leq j \leq K$, and, thus,
$g({\sum_{i=1}^{K}\hat{\bx}_i}/{K})=\rmL_1^*=\cdots=\rmL_K^*$,
which is a contradiction.

Furthermore, a minimization on $\GDd^K$ cannot even be geodesically convex.
Indeed, the maximum of a geodesically convex function on a\vspace*{1pt} compact,
geodesically convex set is attained on the boundary. However, $\GDd^K$
is compact, geodesically convex and has no boundary, so any function
defined on $\GDd^K$ is not geodesically convex.

\subsection{Implications for a single subspace recovery}\label{secsingle}

In \cite{lprecoverypart111}, we discussed the recovery of a single subspace.
Theorems \ref{thmhlm2} and \ref{thmnoisyhlm2} apply to this case when $K=1$.
Unlike \cite{lprecoverypart111} which assumed that $\mu_0$ was
spherically symmetric (while\vspace*{1pt} having possibly additional ``outliers''
along other subspaces, distributed according to $\{\mu_i\}_{i=2}^K$),
here we have a very weak requirement from $\mu_0$ (which represents
all outliers).
However, here there is a strong restriction on the fraction of
outliers, $\alpha_0$, whereas in \cite{lprecoverypart111} there was
no requirement,
except for $\alpha_0 <1$.

\subsection{Extending our theory to more general distributions}
\label{subsecmoregeneral}

In Theorems \ref{thmhlm2} and \ref{thmnoisyhlm2},
the strict spherical symmetry of $\{\mu_i\}_{i=1}^K$ (within
$\{\rmL_i\}_{i=1}^K$, resp.)
can be replaced by approximate spherical symmetry of $\{\mu_i\}_{i=1}^K$.
That is,
for each $1\leq i \leq K$ and $\rmL_i$ and $\mu_i$ as before, we form a new
distribution $\mu_i'$,
with the same support as $\mu_i$ such that the derivative of $\mu_i'$
w.r.t. $\mu_i$
is bounded away from $0$  and $\infty$. We then replace $\mu_i$ with
$\mu_i'$.
This new\vspace*{1pt} setting\vadjust{\goodbreak} will require replacing $\{\alpha_i\}_{i=1}^{K}$
in (\ref{eqcondalpha2})--(\ref{eqfdefine})
by $\{\delta_i \, \alpha_i\}_{i=1}^{K}$,
where $\delta_i \equiv \delta_i(\mu_i',\mu_i)$ for $1 \leq i \leq K$
($\delta_i$ is the lowest value of the derivative
of $\mu_i'$ w.r.t. $\mu_i$).

Furthermore, the boundedness of the support of the distributions $\{\mu
_i\}_{i=0}^K$ can be weakened by assuming that these
distributions are sub-Gaussian. Indeed, this will mainly require
changing Hoeffding's inequality with
\cite{taoranmatbook}, Proposition 2.1.9.

\subsection{Distributions resulting in counterexamples for our theory}\label{seccounterexamples}
There are several typical cases with settings different than above,
where the underlying subspaces cannot be recovered by minimizing the
energy (\ref{eqdeferrorksub}) for all $p>0$.

The first typical example is when there is an outlier with sufficiently
large magnitude so that the minimizer of (\ref{eqdeferrorksub})
contains a subspace passing through this outlier,
which is different than any of the underlying subspaces.
Our setting avoids such a counterexample by requiring (\ref{eqcondalpha2}).
We briefly provide the idea as follows: an arbitrarily large outlier in
our setting of supports within $\ball(\bnull,1)$ means, for example,
that the outlier has magnitude one and the inliers are supported within
$\ball(\bnull,\eps)$, where $\eps$ is arbitrarily small.
Therefore, $\psi(\eps)=1$, so that $\psi_{\mu_1}^{-1} (
{(1+(2K-1)\mu_1(\{\bnull\}))}/{2K})< \psi_{\mu_1}^{-1} (1) = \eps$
and, consequently, $\tau_0 \lessapprox\eps^p$.
In view of (\ref{eqcondalpha2}), we control the fraction of outliers
as a~function of $\eps^p$. In particular,
for a~fixed sample size and sufficiently small~$\eps$, no outliers
are allowed by this condition.

The second example is when the distribution of outliers lies on another
subspace,
$\rmL^*_0 \in\GDd$ and $\alpha_0 > \min_{1 \leq i \leq K} \alpha_i$,
so that $\rmL^*_0$ is contained in the minimizer of (\ref{eqdeferrorksub}).
Our setting avoids this counterexample by assuming an upper bound on
the percentage of outliers in terms of the minimal percentage of inliers
[see (\ref{eqcondalpha2})].

For the last example we assume for simplicity that $D=2$, $d=1$, $K=2$
and underlying uniform distributions
(of outliers and along the two underlying lines) restricted to the unit disk.
We further assume that the two lines have angles $\eps$ and $-\eps$
w.r.t. the $x$-axis.
By choosing $\eps$ sufficiently small the $x$-axis and $y$-axis
provide a smaller value for
the energy (\ref{eqdeferrorksub}) than the underlying lines. We note
that in this case (\ref{eqcondalpha2})
does not hold [due to the small size of $\dG(\rmL^*_i,\rmL^*_j)$].

\subsection{Another phase transition at $p=1$: Many local minima for $0<p<1$}
\label{secp=1}
Our previous work \cite{lprecoverypart111},
proof of Proposition 2.1,
implies that if $0<p<1$ and
there exist distinct subspaces $\{\rmL_i\}_{i=1}^K \subseteq\GDd$
such that
$\Sp(\rmX\cap\rmL_i) =\rmL_i$ for all $1\leq i\leq K$, then $\{
\rmL_i\}_{i=1}^K$
is a local minimizer of the energy (\ref{eqdeferrorksub}). We note
that many subspaces satisfy this condition (in particular,
w.o.p. $d$-subspaces spanned by randomly sampled $d$ vectors).
Therefore, $l_p$ minimization for multiple subspaces with $0<p<1$
will often lead
to plenty of local minima.

This wealth of local minima clearly does not occur when $p=1$ (or $p
\geq1$).
It will be interesting, though difficult, to carefully analyze the
number and depth of local minima for\vadjust{\goodbreak} $p \geq1$.

\subsection{The case of affine subspaces}
\label{secaffinesubs}

Our analysis was restricted to
linear subspaces, though we believe that it can be extended to affine
subspaces. Indeed, we can consider the affine
Grassmannian \cite{Mat95}, which distinguishes between
subspaces according to both their offsets with respect to the origin
(i.e., distances to closest linear subspaces of the same dimension)
and their orientations (based on principal angles of the shifted
linear subspaces). By assuming only affine subspaces intersecting a
fixed ball,
we can have a compact space.
We can also generalize (\ref{eqellprobest}) (with a different function
$\psi_{\mu_1}$)
and the estimates on $\delta_0$ and $\kappa_0$ in Section \ref
{secdeltakappa} to the case of affine subspaces.
We remark, though, that it is not obvious whether the metric on
the affine Grassmannian is relevant for our applications, since it
mixes two different quantities of different units (i.e., offset
values and orientations) so that one can arbitrarily weigh their
contributions. Also, the common strategy of using
homogenous coordinates which transform $d$-dimensional affine
subspaces in~$\reals^D$ to $(d+1)$-dimensional linear subspaces in
$\reals^{D+1}$ is not useful to us since it distorts the structure
of both noise and outliers.

The minimization of the energy (\ref{eqdeferrorksub}) over affine
subspaces seems to
result in more local minima than in the linear case, which can
partially explain why numerical
heuristics for minimizing (\ref{eqdeferrorksub}) do
not perform as well with affine subspaces as they do with linear ones.
We are interested in further
explanation of this phenomenon.

\subsection{The case of mixed dimensions}
\label{secmixeddimensions}

It will be interesting to try to extend our analysis to linear
subspaces of mixed dimensions $d_1, \ldots, d_K$, known in advance.
We believe that it is possible to extend Theorem \ref{thmhlm2} and its
proof to this case.
For this purpose, we suggest using the same distance for subspaces of
the same dimension and
defining the distance $\dG(\rmL_1,\rmL_2)$ between linear subspaces
$\rmL_1$ and $\rmL_2$ of different dimensions (with some abuse of
notation) as follows:
if $\dim(\rmL_1)<\dim(\rmL_2)$, then $\dG(\rmL_1,\rmL_2)=\min
_{\rmL\in\rmL_2, \dim(\rmL)=\dim(\rmL_1)} \dG(\rmL_1,\rmL)$.

\subsection{Further performance guarantees for $l_p$-based HLM algorithms} \label{secotheralg}
We are interested in extending our theory to analyze heuristics (like
the $K$-sub\-spaces)
which try to minimize the $l_p$ energy of (\ref{eqdeferrorksub})
in practice.


\subsection{Asymptotic rates of convergence and sample complexity}

In Section~\ref{secfapproach0} we demonstrated simple instances when
noise is present and
one cannot asymptotically recover the underlying
subspaces by $l_p$ \mbox{minimization} for all $p>0$.
One may still inquire about the existence of asymptotic limit different
than the underlying subspaces and quantify the rate of convergence
(depending on
the mixture model parameters) to that limit.
That is, assume that $\{\hat{\rmL}_1, \hat{\rmL}_2\}$ is the
minimizer of
$\shrtexp_{\mu}(l_p(\bx,\rmL_1,\rmL_2))$ and $\{\hat
{\rmL}_1^N, \hat{\rmL}_2^N\}$
is the minimizer of $\shrtexp_{\mu_N}(l_p(\bx,\rmL_1,\rmL
_2))$, where $\mu_N$ is an
empirical distribution of i.i.d. sample of $N$ points from $\mu$.
We first ask whether \mbox{$\dist(\{\hat{\rmL}_1, \hat{\rmL}_2\},\{\hat
{\rmL}_1^N, \hat{\rmL}_2^N\})\rightarrow0$}\vadjust{\goodbreak}
as $N\rightarrow\infty$. If true, then we ask about the asymptotic
rates of convergence. This will then allow a definition of a sample complexity
for multiple subspaces as the number of samples required to achieve a
prediction error within~$\varepsilon$ of the exact recovery of the $K$ $d$-subspaces.

\begin{appendix}\label{app}
\section*{Appendix: Supplementary details}
\subsection{\texorpdfstring{Proof of Lemma \protect\ref{lemmaelldistest}}{Proof of Lemma 2.1}}
\label{applemmaproof2}

We will use the following inequality for any $1\leq j\leq K$, which is
proved in \cite{lprecoverypart111}, Section A.1.1:
%
%
\begin{eqnarray}\label{eqellprobest}
&&\mu_1 \bigl(\bx\in
\ball(\mathbf{0},1)\cap\rmL^*_1\dvtx
\dist(\bx,\hat{\rmL}_j)<\beta\dG(\rmL^*_1,\hat{\rmL
}_j)\bigr)\nonumber\\[-8pt]\\[-8pt]
&&\qquad\leq
\psi_{\mu_1}\biggl(\frac{\pi\sqrt{d}}{2}\beta\biggr) \qquad\forall
\beta>0.\nonumber
\end{eqnarray}
%
We denote $\beta_1\,{=}\,\frac{2}{\pi\sqrt{d}}\psi_{\mu_1}^{-1}
(\frac{1+(2K-1)\mu_1(\{\bnull\})}{2K})$ (the existence of
$\psi_{\mu_1}^{-1}(\frac{1+(2K-1)\mu_1(\{\bnull\}
)}{2K})$ follows the same proof as in \cite{lprecoverypart111},
Section A.1.1) and
combine (\ref{eqellprobest}) with the fact that $\dG(\rmL^*_1,\hat
{\rmL}_j)\geq\eps$ for any $1\leq j\leq K$ to obtain that
\begin{eqnarray*}
&&\mu_1\bigl(\bx\in\ball(\mathbf{0},1)\cap\rmL^*_1\setminus\{
\bnull\}\dvtx\dist(\bx,\hat{\rmL}_1)<\beta_1 \eps\bigr)\\
&&\qquad= \mu_1\bigl(\bx\in\ball(\mathbf{0},1)\cap\rmL^*_1\setminus\{
\bnull\}\dvtx
\dist(\bx,\hat{\rmL}_1)<\beta_1 \dG(\rmL^*_1,\hat{\rmL
}_1)\bigr)\\
&&\qquad\leq
\frac{1+(2K-1)\mu_1(\{\bnull\})}{2K}-\mu(\{\bnull\})\\
&&\qquad=\frac{1-\mu
_1(\{\bnull\})}{2K}.
\end{eqnarray*}
Consequently,
\begin{eqnarray*}
&&\mu_1\Biggl(\bx\in\ball(\mathbf{0},1)\cap\rmL^*_1\dvtx\dist\Biggl(\bx
,\bigcup_{j=1}^{K}\hat{\rmL}_1\Biggr)\geq\beta_1\eps\Biggr)\\
&&\qquad\geq 1-\mu(\{\bnull\})
- \sum_{i=1}^{K}\mu_1\bigl(\bx\in\ball
(\mathbf{0},1)\cap\rmL^*_1\setminus\{\bnull\}\dvtx\dist(\bx,\hat
{\rmL}_i)<\beta_1 \eps\bigr)\\
&&\qquad\geq\bigl(1-\mu_1(\{\bnull\})\bigr)/2,
\end{eqnarray*}
and, thus, by Chebyshev's inequality the lemma is concluded as
follows:
\begin{eqnarray*}
\shrtexp_{{\mu}_1}(e_{l_p}(\bx,\hat{\rmL}_1))&\geq&
\beta_1^p\eps^p/2\\
&=&\frac{(1-\mu_1(\{\bnull\}))2^{p-1}\psi_{\mu
_1}^{-1}(({1+(2K-1)\mu_1(\{\bnull\})})/({2K}))^p\eps
^p}{(\pi\sqrt{d})^p}\\
&=&\tau_0 \eps^p.
\end{eqnarray*}

\subsection{\texorpdfstring{Proof of Proposition \protect\ref{proprmYibest}}{Proof of Proposition 3.1}}
\label{apprmYibest}

The proof is an immediate consequence of the following inequality,
which uses an arbitrary ${\rmL}_1\in\GDd$ and the\vadjust{\goodbreak} notation
$\rmY'_i = \rmY_i(\rmL'_1,\ldots,\rmL'_K)$, $1 \leq i \leq K$:
%
\begin{eqnarray*}
0&\leq&\shrtexp_{\nu}(e_{l_p}(\bx,
{\rmL}_1,{\rmL}'_2,\ldots,{\rmL}'_K))-\shrtexp_{\nu}(e_{l
_p}(\bx,
{\rmL}'_1,\ldots,{\rmL}'_K))\\
&\leq&\shrtexp_{\nu}\bigl(I(\bx\in\rmY'_1)e_{l_p}(\bx,
{\rmL}_1)\bigr) + \sum_{2\leq i\leq K}\shrtexp_{\nu}\bigl(I(\bx\in
\rmY'_i)e_{l_p}(\bx, {\rmL}'_i)\bigr)\\
&&{}-\sum_{1\leq i\leq
K}\shrtexp_{\nu}\bigl(I(\bx\in\rmY'_i)e_{l_p}(\bx,
{\rmL}'_i)\bigr)\\
&=&\shrtexp_{\nu}\bigl(I(\bx\in\rmY'_1)e_{l_p}(\bx,
{\rmL}_1)\bigr)-\shrtexp_{\nu}\bigl(I(\bx\in
\rmY'_1)e_{l_p}(\bx, {\rmL}'_1)\bigr).
\end{eqnarray*}

\subsection{\texorpdfstring{Proof of Lemma \protect\ref{lemmaintersection}: Geometric sensitivity}
{Proof of Lemma 4.2: Geometric sensitivity}}
\label{secgeosensitive}
We will first show that there exists $\bx_0 \in\ball(\bnull,1)$
such that
%
%
\begin{equation}\label{eqbxnearest}
\dist(\bx_0,\rmL^*_1)=\dist(\bx_0,\rmL^*_2)<\min_{3\leq i\leq
K}\dist(\bx_0,\rmL^*_i).
\end{equation}
We verify (\ref{eqbxnearest}) in two cases: $d^*=d$ and $d^*=D-d$.
We will then prove that~(\ref{eqbxnearest}) implies (\ref{eqintersection}).
Throughout the proof we denote the principal vectors of~$\rmL^*_2$ and
$\rmL^*_1$ by
$\{\hat{\mathbf{v}}_i\}_{i=1}^{d^*}$ and $\{\mathbf{v}_i\}
_{i=1}^{{d^*}}$, respectively.

\subsubsection{\texorpdfstring{Part \textup{I}: Proof of (\protect\ref{eqbxnearest}) when $d^*=d$}
{Part \textup{I}: Proof of (71) when $d^*=d$}}
\label{secpart1bxnearest}
We define
\[
\bx_0=(\hat{\mathbf{v}}_{d^*}+\mathbf{v}_{d^*})/{\|\hat{\mathbf
{v}}_{d^*}+\mathbf{v}_{d^*}\|}
\]
and arbitrarily fix $i_0 >3$ and $\mathbf{v}_0 \in\rmL^*_{i_0}$. We
will show that
%
%
\begin{equation}
\label{eqlemsenpart1}
\ang(\bx_0,\mathbf{v}_0)>\theta_{d^*}(\rmL^*_2,\rmL^*_1)/2
\end{equation}
and consequently conclude (\ref{eqbxnearest}) as follows:
\begin{eqnarray*}
\dist(\bx_0,\rmL^*_{i_0} ) &\geq& \sin(\ang(\bx_0,\mathbf
{v}_0)) > \sin\bigl(\theta_{d^*} (\rmL^*_2,\rmL^*_1)/2\bigr) = \dist
(\bx_0,\rmL^*_1) \\
&=& \dist(\bx_0,\rmL^*_2).
\end{eqnarray*}

We can easily verify a weaker version of (\ref{eqlemsenpart1}) where
the inequality is not necessarily strict.
Indeed, using elementary geometric estimates and the fact that the
intersections of the $d$-subspaces $\{\rmL^*_i\}_{i=1}^K$ are empty
[which follows
from (\ref{eqcondition0})], we obtain that
%
%
\begin{eqnarray}\label{eqbuv}\qquad
\ang(\bx_0,\mathbf{v}_0)&\,{\geq}\,&\ang(\mathbf{v}_{d^*},\mathbf
{v}_0)\,{-}\,\ang(\mathbf{v}_{d^*},\bx_0)\,{\geq}\,\theta_{d^*}(\rmL
^*_{i_0},\rmL^*_1)\,{-}\,\theta_{d^*}(\rmL^*_2,\rmL^*_1)/2\nonumber\\[-8pt]\\[-8pt]
&\,{\geq}\,&\theta_{d^*}(\rmL^*_2,\rmL^*_1)\,{-}\,\theta_{d^*}(\rmL^*_2,\rmL
^*_1)/2\,{=}\,\theta_{d^*}(\rmL^*_2,\rmL^*_1)/2.\nonumber
\end{eqnarray}

At last, we show that (\ref{eqbuv}) cannot be an equality. Indeed, if
the first inequality in (\ref{eqbuv}) is an equality, then $\mathbf
{v}_0$, $\mathbf{v}_{d^*}$ and $\bx_0$ are on a geodesic line within
the sphere $S^{D-1}$.
Combining this with the assumption that all other inequalities
in (\ref{eqbuv}) are equalities, we obtain that
$\ang(\bx_0,\mathbf{v}_0)=\theta_{d^*}(\rmL^*_2,\rmL^*_1)/2=\ang
(\bx_0,\mathbf{v}_{d^*})=\ang(\bx_0,\hat{\mathbf{v}}_{d^*})$.
This implies that
either $\mathbf{v}_0=\hat{\mathbf{v}}_{d^*}$ or $\mathbf
{v}_0=\mathbf{v}_{d^*}$, which contradicts (\ref{eqcondition0}).

\subsubsection{\texorpdfstring{Part \textup{II}: Proof of (\protect\ref{eqbxnearest}) when $d^*=D-d$}
{Part \textup{II}: Proof of (71) when $d^*=D-d$}}
It follows from basic dimension equalities of subspaces and (\ref
{eqcondition0}) that for all\vadjust{\goodbreak} $2 \leq i \leq K\dvtx\dim(\rmL^*_1 \cup
\rmL^*_i)=D$ and $\dim(\rmL^*_1 \cap\rmL^*_i)=2d-D$.
We denote by $K_0$ the integer in $\{0,\ldots,K\}$ such that
for any $3\leq i \leq K_0\dvtx\rmL^*_1\cap\rmL^*_i=\rmL^*_1\cap\rmL
^*_2$ and for any $i>K_0\dvtx\rmL^*_1\cap\rmL^*_i\neq\rmL^*_1\cap
\rmL^*_2$
(the existence of $K_0$ may require reordering of the indices of the
subspaces $\{\rmL^*_i\}_{i=3}^K$).
In order to define $\bx_0$ in the current case, we let $\bx_1=(\hat
{\mathbf{v}}_{d^*}+\mathbf{v}_{d^*})/{\|\hat{\mathbf
{v}}_{d^*}+\mathbf{v}_{d^*}\|}$,
$\bx_2$ be an arbitrarily fixed unit vector in $\rmL^*_1\cap(\rmL
^*_2\setminus\bigcup_{K_0< i\leq K}\rmL^*_i)$, $\eps_0 = \dist(\bx
_2,\bigcup_{K_0< i\leq K}\rmL^*_i)$ and
\[
\bx_0=\bx_2/2+\eps_0 \bx_1/5.
\]

We first claim that
%
%
\begin{equation}\label{eqbxleqK0}
\dist(\bx_0,\rmL^*_1)=\dist(\bx_0,\rmL^*_2)<\min
_{3\leq j \leq K_0}\dist(\bx_0,\rmL^*_j).
\end{equation}
Indeed, we can remove $\rmL^*_1 \cap\rmL^*_2$ from the subspaces $\{
\rmL^*_i\}_{i=1}^{K_0}$ and obtain subspaces of dimension $D-d$
intersecting each other at the origin. We can then rewrite (\ref
{eqbxleqK0}) by replacing $\{\rmL^*_i\}_{i=1}^{K_0}$ with their reduced
version and $\bx_0$ with~$\bx_1$. The argument of Section \ref
{secpart1bxnearest} thus proves this equation.

We conclude (\ref{eqbxnearest}) by combining (\ref{eqbxleqK0}) with
the following observation:
%
%
\begin{eqnarray}\label{eqbx>K0}\quad
\dist(\bx_0,\rmL^*_1)&=&\eps_0 \dist(\bx_1,\rmL^*_1)/5
\leq\eps_0/5 < \dist\biggl(\bx_2/2,\bigcup_{K_0<
j \leq K}\rmL^*_j\biggr)-\eps_0/5\nonumber\\[-8pt]\\[-8pt]
&\leq&\dist\biggl(\bx_2/2+\eps_0 \bx_1/5,\bigcup_{K_0<
j \leq K}\rmL^*_j\biggr)
=
\min_{K_0<
i \leq K}\dist(\bx_0,\rmL^*_i).
\nonumber
\end{eqnarray}
%

\subsubsection{\texorpdfstring{Part \textup{III}: Deriving (\protect\ref{eqintersection})
from (\protect\ref{eqbxnearest}) in a simple case}
{Part \textup{III}: Deriving (47) from (71) in a simple case}}

We note that~(\ref{eqbxnearest}) implies that
%
%
\begin{equation}\label{eqboundbx0}
\bx_0\in\bigl(\rmY_1\cup\rmY_2\cup(\bar{\rmY}_1\cap\bar{\rmY}_2)\bigr)
\cap
\bigl(\hat{\rmY}_1\cup\hat{\rmY}_2\cup(\bar{\hat{\rmY}}_1\cap\bar
{\hat{\rmY}}_2)\bigr)
\end{equation}
and, consequently,
%
%
\begin{equation}\label{eqboundbx}
\ball(\bx_0,\eps)\subset
\bigl(\rmY_1\cup\rmY_2\cup(\bar{\rmY}_1\cap\bar{\rmY}_2)\bigr) \cap
\bigl(\hat{\rmY}_1\cup\hat{\rmY}_2\cup(\bar{\hat{\rmY}}_1\cap\bar
{\hat{\rmY}}_2)\bigr).
\end{equation}
We will deduce here (\ref{eqintersection}) from (\ref{eqboundbx}) in
the simpler case:
$\bar{\hat{\rmY}}_1\cap\bar{\hat{\rmY}}_2\cap\ball(\bx_0,\eps
)\neq\bar{\rmY}_1\cap\bar{\rmY}_2\cap\ball(\bx_0,\eps)$.

Using (\ref{eqboundbx}) and the fact that $\calL_D(\bar{\rmY}_1\cap
\bar{\rmY}_2)=0$, we may choose
$\by\in(\bar{\hat{\rmY}}_1\cap\bar{\hat{\rmY}}_2\cap\ball
(\bx_0,\eps)) \cap({\rmY}_1 \cup{\rmY}_2)$; WLOG we assume
instead of the latter condition that $\by\in(\bar{\hat{\rmY
}}_1\cap\bar{\hat{\rmY}}_2\cap\ball(\bx_0,\eps)) \cap{\rmY}_1$.
By slightly\vspace*{1pt} perturbing $\by$ we can choose another point $\by_0$ such
that $\by_0\in{\hat{\rmY}}_2$ and $\by_0\in{\rmY}_1\setminus
\hat{\rmY}_1$. It follows from the continuity of the distance
function that there exists a small $\eta>0$ such that $(\hat{\rmY
}_1\setminus\rmY_1)\cup({\rmY}_1\setminus\hat{\rmY}_1)\supseteq
{\rmY}_1\setminus\hat{\rmY}_1\supset\ball(\by_0,\eta)$, which
proves (\ref{eqintersection}).

\subsubsection{\texorpdfstring{Part \textup{IV}: Deriving (\protect\ref{eqintersection})
from (\protect\ref{eqbxnearest}) in the complementary case}
{Part \textup{IV}: Deriving (47) from (71) in the complementary case}}

At\vspace*{1pt} last, we assume that $\bar{\hat{\rmY}}_1\cap\bar{\hat{\rmY
}}_2\cap\ball(\bx_0,\eps)= \bar{\rmY}_1\cap\bar{\rmY}_2\cap
\ball(\bx_0,\eps)$. We show here that it leads to the
contradiction:\vadjust{\goodbreak}
$\hat{\rmL}_2=\rmL^*_2$.

We note that the sets of solutions in $\ball(\bx_0,\eps)$ of the
equations $\bx^T(P_{\rmL^*_1}-P_{\rmL^*_2})\bx=0$ and $\bx
^T(P_{\rmL^*_1}-P_{\hat{\rmL}_2})\bx=0$
are\vspace*{1pt} $\bar{\hat{\rmY}}_1\cap\bar{\hat{\rmY}}_2\cap\ball(\bx
_0,\eps)$ and $\bar{\rmY}_1\cap\bar{\rmY}_2\cap\ball(\bx
_0,\eps)$, respectively.
In view of (\ref{eqboundbx}), these solution sets coincide. They are
$(D-1)$-manifolds
and, thus, their $(D-1)$-dimensional tangent spaces at~$\bx_0$, that
is, $\bx_0^T (P_{\rmL^*_1}-P_{\rmL^*_2}) = \bnull$ and $\bx_0^T
(P_{\rmL^*_1}-P_{\hat{\rmL}_2})=\bnull$, also coincide.
Consequently, we have that $\bx_0^T(P_{\rmL^*_1}-P_{\rmL^*_2})=t_0
\bx_0^T(P_{\rmL^*_1}-P_{\hat{\rmL}_2})$ for some \mbox{$t_0\neq0$}.
Similarly, for any $\bx_1\in\bar{\hat{\rmY}}_1\cap\bar{\hat
{\rmY}}_2\cap\ball(\bx_0,\eps)$, we have $\bx_1^T(P_{\rmL
^*_1}-P_{\rmL^*_2})=t_1 \bx_1^T(P_{\rmL^*_1}-P_{\hat{\rmL}_2})$
for some $t_1\neq0$. We note that $t_1=t_0$ by the following argument:
$t_1 \bx_1^T(P_{\rmL^*_1}-P_{\hat{\rmL}_2})\bx_0=\bx_1^T(P_{\rmL
^*_1}-P_{\rmL^*_2})\bx_0=t_0 \bx_1^T(P_{\rmL^*_1}-P_{\hat{\rmL
}_2})\bx_0$.
Therefore, there exists $t \neq0$ such that for any $\bx_1\in\bar
{\hat{\rmY}}_1\cap\bar{\hat{\rmY}}_2\cap\ball(\bx_0,\eps)$,
%
%
\begin{equation}\label{eqtangent}
\bx_1^T(P_{\rmL^*_1}-P_{\rmL^*_2})=t \bx_1^T(P_{\rmL^*_1}-P_{\hat
{\rmL}_2}).
\end{equation}

Since the tangent space of $\bar{\hat{\rmY}}_1\cap\bar{\hat{\rmY
}}_2\cap\ball(\bx_0,\eps)$ [or, equivalently, $\bx^T(P_{\rmL
^*_1}-P_{\hat{\rmL}_2})\bx=0$] at $\bx_0$ has dimension $D-1$, the
subspace $\rmL^*_0=
\Sp(\bar{\hat{\rmY}}_1\cap\bar{\hat{\rmY}}_2\cap\ball(\bx
_0,\eps))$ [i.e., the closure of all finite linear combinations of
vectors in $\bar{\hat{\rmY}}_1\cap\bar{\hat{\rmY}}_2\cap\ball
(\bx_0,\eps)$] has dimension at least $D-1$.
In view of (\ref{eqtangent}), $\rmL^*_0$ satisfies
%
%
\begin{equation}\label{eqrmL0}
P_{\rmL^*_0}(P_{\rmL^*_1}-P_{\rmL^*_2})=t P_{\rmL^*_0}(P_{\rmL
^*_1}-P_{\hat{\rmL}_2}).
\end{equation}
Due to the symmetry of $(P_{\rmL^*_1}-P_{\hat{\rmL}_2})$ and
$(P_{\rmL^*_1}-P_{\rmL^*_2})$,
we have the following equivalent formulation of (\ref{eqrmL0}):
%
%
%
\begin{equation}\label{eqrml01}
(P_{\rmL^*_1}-P_{\rmL^*_2})P_{\rmL^*_0} =
(P_{\rmL^*_1}-P_{\hat{\rmL}_2}) P_{\rmL^*_0}.
\end{equation}
Furthermore, using the fact that $(P_{\rmL^*_1}-P_{\hat{\rmL}_2})$
and $(P_{\rmL^*_1}-P_{\rmL^*_2})$ have trace $0$, we obtain that
%
%
\begin{eqnarray}\label{eqrml02}
\tr\bigl(P_{\rmL^{*\perp}_1}(P_{\rmL^*_1}-P_{\rmL^*_2})P_{\rmL^{*\perp}_0}\bigr)&=&
-\tr\bigl(P_{\rmL^*_0}(P_{\rmL^*_1}-P_{\rmL^*_2})P_{\rmL^*_0}\bigr)
\nonumber\\
&=& -t \cdot\tr\bigl(P_{\rmL^*_0}(P_{\rmL^*_1}-P_{\hat{\rmL
}_2})P_{\rmL^*_0}\bigr)\\
&=&t\cdot\tr\bigl(P_{\rmL^{*\perp}_0}(P_{\rmL^*_1}-P_{\hat{\rmL
}_2})P_{\rmL^{*\perp}_0}\bigr).
\nonumber
\end{eqnarray}
Since $P_{\rmL^{*\perp}_0}$ is at most one-dimensional,
(\ref{eqrml02}) can be rewritten as
%
%
\begin{equation}\label{eqrml03}
P_{\rmL^{*\perp}_0}(P_{\rmL^*_1}-P_{\rmL^*_2})P_{\rmL^{*\perp}_0}=
t\cdot\bigl(P_{\rmL^{*\perp}_0}(P_{\rmL^*_1}-P_{\hat{\rmL}_2})P_{\rmL
^{*\perp}_0}\bigr).
\end{equation}
Combining (\ref{eqrmL0}), (\ref{eqrml01}) and (\ref{eqrml03}), we
obtain that $(P_{\rmL^*_1}-P_{\hat{\rmL}_2})=t (P_{\rmL
^*_1}-P_{\rmL^*_2})$, equivalently,
%
%
\begin{equation}
\label{eqformulapl2}
P_{\hat{\rmL}_2}=(1-t) P_{\rmL^*_1}+t P_{\rmL^*_2}.
\end{equation}

We conclude the desired contradiction in two different cases. Assume
first that $t< 1$ and let $\mathbf{v}_0$ be an arbitrary unit vector
in $\rmL^*_2$. We note that $\mathbf{v}_0^TP_{\hat{\rmL}_2}\mathbf
{v}_0=1$ as well as $(1-t) \mathbf{v}_0^TP_{\rmL^*_1}\mathbf{v}_0=
1- t \mathbf{v}_0^TP_{\rmL^*_2}\mathbf{v}_0\geq1-t$. Consequently,
$\mathbf{v}_0^TP_{\rmL^*_1}\mathbf{v}_0=1$, that is, $\mathbf
{v}_0\in\rmL^*_1$ and, thus, we obtain the following contradiction
with (\ref{eqcondition0}):
$\rmL^*_1=\hat{\rmL}_2$ [in view of (\ref{eqformulapl2}), this is
equivalent with $\hat{\rmL}_2=\rmL^*_2$]. Next, assume that $t \geq
1$ and, as before, $\mathbf{v}_0$ is an arbitrary unit vector in~$\rmL
^{*\perp}_2$. In this case, $\mathbf{v}_0^TP_{\hat{\rmL}_2}\mathbf
{v}_0=(1-t) \mathbf{v}_0^TP_{\rmL^*_1}\mathbf{v}_0+t \mathbf
{v}_0^TP_{\rmL^*_2}\mathbf{v}_0 \leq0+0=0$. Therefore, $\mathbf
{v}_0\in\hat{\rmL}_2^\perp$ and
we obtain the following contradiction with (\ref{eqcondition0}): $\rmL
^*_2=\hat{\rmL}_2$. Equation (\ref{eqintersection}) is thus proved.
\end{appendix}

\section*{Acknowledgments}

Our collaboration with Arthur Szlam on
efficient and fast algorithms for hybrid linear modeling (especially via
geometric $l_1$ minimization) inspired this investigation. We thank
John Wright for interesting discussions and J.~Tyler Whitehouse
for commenting on an earlier version of this manuscript. Thanks to
the Institute for Mathematics and its Applications (IMA), in
particular, Doug Arnold and Fadil Santosa, for holding a~workshop on
multi-manifold modeling that G. Lerman co-organized and T. Zhang
participated in.
G. Lerman thanks David
Donoho for inviting him for a visit to Stanford University in Fall
2003 and for stimulating discussions at that time on the
intellectual responsibilities of mathematicians analyzing massive
and high-dimensional data as well as general advice. Those
discussions effected G. Lerman's research program
and his mentorship (T. Zhang is
a Ph.D. candidate advised by G. Lerman).


%

\printaddresses

\end{document}